\documentclass[10pt,journal,letterpaper,compsoc]{IEEEtran}

\newif\ifarxiv
\arxivtrue
\usepackage{times}
\usepackage{natbib}
\usepackage{graphicx} 
\usepackage{afterpage}
\usepackage{amsmath}
\usepackage{amssymb}
\usepackage{color}
\usepackage[utf8]{inputenc}
\usepackage{refcount}

\def\data{{\rm data}}
\def\energy{{\cal E}}
\def\sigmoid{{\rm sigmoid}}
\def\note#1{}
\def\begplan{\color{red}}
\def\endplan{\color{black}}

\def\vsA{\vspace*{-0.5mm}}
\def\vsB{\vspace*{-1mm}}
\def\vsC{\vspace*{-2mm}}
\def\vsD{\vspace*{-3mm}}
\def\vsE{\vspace*{-4mm}}
\def\vsF{\vspace*{-5mm}}
\def\vsG{\vspace*{-6mm}}

\newcommand{\argmin}{\operatornamewithlimits{argmin}}
\newcommand{\argmax}{\operatornamewithlimits{argmax}}
\newcommand{\E}[2]{ {\mathbb{E}}_{#1}\left[{#2}\right] }
\newcommand{\EE}[1]{ {\mathbb{E}}\left[{#1}\right] }
\newcommand{\R}{ {\mathbb{R}} }

\begin{document}

\title{Representation Learning: A Review and New Perspectives}

\author{Yoshua Bengio$^\dagger$,
        Aaron Courville,
        and Pascal Vincent$^\dagger$\\
Department of computer science and operations research, U. Montreal\\
$\dagger$ also, Canadian Institute for Advanced Research (CIFAR)
\vsC
\vsC
\vsC
}
\date{}
\maketitle
\vsC
\vsC
\vsC
\begin{abstract}
\vsA

The success of machine learning algorithms generally depends on data representation, and we hypothesize that this is because different representations can entangle and hide more or less the different explanatory factors of variation behind the data.  Although specific domain knowledge can be used to help design representations, learning with generic priors can also be used, and the quest for AI is motivating the design of more powerful representation-learning algorithms implementing such priors. This paper reviews recent work in the area of unsupervised feature learning and deep learning, covering advances in probabilistic models, auto-encoders, manifold learning, and deep networks.  This motivates longer-term unanswered questions about the appropriate objectives for learning good representations, for computing representations (i.e., inference), and the geometrical connections between representation learning, density estimation and manifold learning.

\end{abstract}
\vsC


\vsC

\vsD
\section{Introduction \note{YB}}
\vsA

The performance of machine learning methods is heavily dependent on the
choice of data representation (or features) on which they are applied.  For
that reason, much of the actual effort in deploying machine learning
algorithms goes into the design of preprocessing pipelines and data
transformations that result in a representation of the data that can
support effective machine learning.  Such feature engineering is important
but labor-intensive and highlights the weakness of current learning
algorithms: their inability to extract and organize the discriminative
information from the data.  Feature engineering is a way to take advantage
of human ingenuity and prior knowledge to compensate for that weakness. In
order to expand the scope and ease of applicability of machine learning, it
would be highly desirable to make learning algorithms less dependent on
feature engineering, so that novel applications could be constructed
faster, and more importantly, to make progress towards Artificial
Intelligence (AI). An AI must fundamentally {\em understand the world
  around us}, and we argue that this can only be achieved if it can learn
to identify and disentangle the underlying explanatory factors hidden in
the observed milieu of low-level sensory data. 

This paper is about {\em representation learning},
i.e., learning representations of the data that make it easier to extract
useful information when building classifiers or other predictors. In the
case of probabilistic models, a good representation is often one that
captures the posterior distribution of the underlying explanatory factors
for the observed input.  A good representation
is also one that is useful as input to a supervised predictor.
Among the various ways of learning
representations, this paper focuses on deep learning methods: those that
are formed by the composition of multiple non-linear transformations,
with the goal of yielding more abstract -- and ultimately more useful
-- representations.
Here we survey this rapidly developing area with special emphasis on recent
progress.  We consider some of the fundamental questions that have been
driving research in this area. Specifically, what makes one representation
better than another? Given an example, how should we compute its
representation, i.e. perform feature extraction? Also, what are appropriate
objectives for learning good representations?

\note{AC: This paragraph bellow doesn't seem to add to much other than a huge list of questions? what role does it serve?}
\note{YB: I like questions. I think they are important. They set the stage. They
raise the issues. They fit well in an introduction.}
\note{AC: This list reads like you wrote it for you, not our target reader. You introduce ideas you don't explain (e.g. explaining away) and almost all of it way too briefly discussed for anyone who isn't an insider to get anything out of it. I have integrated some of these ideas in the above paragraph.}

\vsD
\section{Why should we care about learning representations?}
\label{sec:motivation}
\vsA

Representation learning has become a field in itself in the machine
learning community, with regular workshops at the leading conferences such
as NIPS and ICML, and a new conference dedicated to it, ICLR\footnote{International
Conference on Learning Representations}, sometimes under the header of {\em Deep Learning} or {\em
  Feature Learning}. Although depth is an important part of the story, many
other priors are interesting and can be conveniently captured 
when the problem is cast as one of learning a representation, as
discussed in the next section. The rapid increase in scientific activity
on representation learning has been accompanied and nourished 
by a remarkable
string of empirical successes both in academia and in
industry. Below, we briefly highlight some of these high points.

\vspace*{1mm}
\noindent{\bf Speech Recognition and Signal Processing}
\vspace*{1mm}

Speech was one of the early applications of neural networks,
in particular convolutional (or time-delay) neural networks
\footnote{See~\citet{Bengio-ijprai93} for a review of early work in this
  area.}. The recent revival of interest in neural networks,
deep learning, and representation learning has had
a strong impact in the area of speech recognition,
with breakthrough 
results~\citep{dahl2010phonerec-small,Deng-2010,Seide2011,Mohamed+Dahl+Hinton-2012,Dahl2012,Hinton-et-al-2012} 
obtained by several academics as well as researchers at
industrial labs bringing these
algorithms to a larger scale and into products. For example, Microsoft
has released in 2012 a new version of their MAVIS (Microsoft Audio Video Indexing Service) 
speech system based on deep learning~\citep{Seide2011}.
These authors managed to reduce the word error rate on 
four major benchmarks by about 30\% (e.g. from 27.4\%
to 18.5\% on RT03S) compared to state-of-the-art models
based on Gaussian mixtures for the acoustic modeling
and trained on the same amount of data (309 hours of speech).
The relative improvement in error rate obtained by
\citet{Dahl2012} on a smaller large-vocabulary speech recognition
benchmark (Bing mobile business search dataset, with 40 hours of speech) 
is between 16\% and 23\%.

Representation-learning algorithms
have also been applied to music, substantially beating the state-of-the-art in polyphonic
transcription~\citep{Boulanger+al-ICML2012-small}, with relative error improvement
between 5\% and 30\% on a standard benchmark of 4 datasets.
Deep learning also helped to win MIREX (Music Information Retrieval)
competitions, e.g. in 2011 on audio tagging~\citep{Hamel-et-al-ISMIR2011-small}.

\vspace*{1mm}
\noindent{\bf Object Recognition}
\vspace*{1mm}

The beginnings of deep learning in 2006 have focused on the MNIST digit
image classification problem~\citep{Hinton06,Bengio-nips-2006-small},
breaking the supremacy of SVMs (1.4\% error) on this dataset\footnote{for the knowledge-free
version of the task, where no image-specific prior is used, such as image
deformations or convolutions}. 
The latest records
are still held by deep networks:
~\citet{Ciresan-2012}  currently claims the title of state-of-the-art for the unconstrained version of the task
(e.g., using a convolutional architecture), with 0.27\% error,
and~\citet{Dauphin-et-al-NIPS2011-small} is state-of-the-art for the
knowledge-free version of MNIST, with 0.81\% error.

In the last few years, deep learning has moved from digits to object recognition in 
natural images, and the latest breakthrough has been achieved
on the ImageNet dataset\footnote{The 1000-class ImageNet benchmark, 
whose results are detailed here:\\
{\tt\scriptsize http://www.image-net.org/challenges/LSVRC/2012/results.html}}
bringing down the state-of-the-art error rate from 26.1\% to
15.3\%~\citep{Krizhevsky-2012-small}.

\vspace*{1mm}
\noindent{\bf Natural Language Processing}
\vspace*{1mm}

Besides speech recognition, there are many other Natural Language Processing (NLP)
applications of representation learning. {\em Distributed
representations} for symbolic data were introduced by ~\citet{Hinton86b-small}, 
and first developed in the context of statistical language modeling 
by~\citet{Bengio-nnlm2003-small} in so-called {\em neural net language
models}~\citep{Bengio-scholarpedia-2007-small}. They are all based on learning
a distributed representation for each word, called a {\em word embedding}.
Adding a convolutional architecture, ~\citet{collobert:2011b} developed the
SENNA system\footnote{downloadable from {\tt http://ml.nec-labs.com/senna/}}
that shares representations across the tasks of language modeling,
part-of-speech tagging,
chunking, named entity recognition, semantic role labeling and syntactic
parsing.  SENNA approaches or surpasses the state-of-the-art on
these tasks but is simpler and much faster than traditional predictors. 
Learning word embeddings can be combined with learning image representations
in a way that allow to associate text and images. This approach
has been used successfully to build Google's image search, exploiting
huge quantities of data to map images and queries in the same space~\citep{Weston+Bengio+Usunier-2010}
and it has recently been extended to deeper 
multi-modal representations~\citep{Srivastava+Salakhutdinov-NIPS2012-small}.

The neural net language model was also improved by adding recurrence to the
hidden layers~\citep{Mikolov-Interspeech-2011-small}, allowing it to beat the state-of-the-art (smoothed n-gram models) not only in terms of perplexity (exponential of the average negative
log-likelihood of predicting the right next word, going down from 140 to 102) 
but also in terms of word error
rate in speech recognition (since the language model is an important component
of a speech recognition system), decreasing it from 17.2\% (KN5 baseline)
or 16.9\% (discriminative language model) to 14.4\% on the Wall Street Journal
benchmark task. Similar models have been applied in statistical
machine translation~\citep{Schwenk-2012,Le-et-al-2013-small}, improving perplexity
and BLEU scores. Recursive auto-encoders (which generalize recurrent
networks) have also been used to beat the state-of-the-art in
full sentence paraphrase detection~\citep{Socher+al-NIPS2011}
almost doubling the F1 score for paraphrase detection.
Representation learning can also be used to perform word sense
disambiguation~\citep{Antoine-al-2012-small}, bringing up the
accuracy from 67.8\% to 70.2\% on the subset of Senseval-3 where
the system could be applied (with subject-verb-object sentences).
Finally, it has also been successfully used to surpass the
state-of-the-art in sentiment analysis~\citep{Glorot+al-ICML-2011-small,Socher+al-EMNLP2011-small}.

\vspace*{1mm}
\noindent{\bf Multi-Task and Transfer Learning, Domain Adaptation}
\vspace*{1mm}

\begin{figure}[h]
\vsC
\centerline{\includegraphics[width=0.6\linewidth]{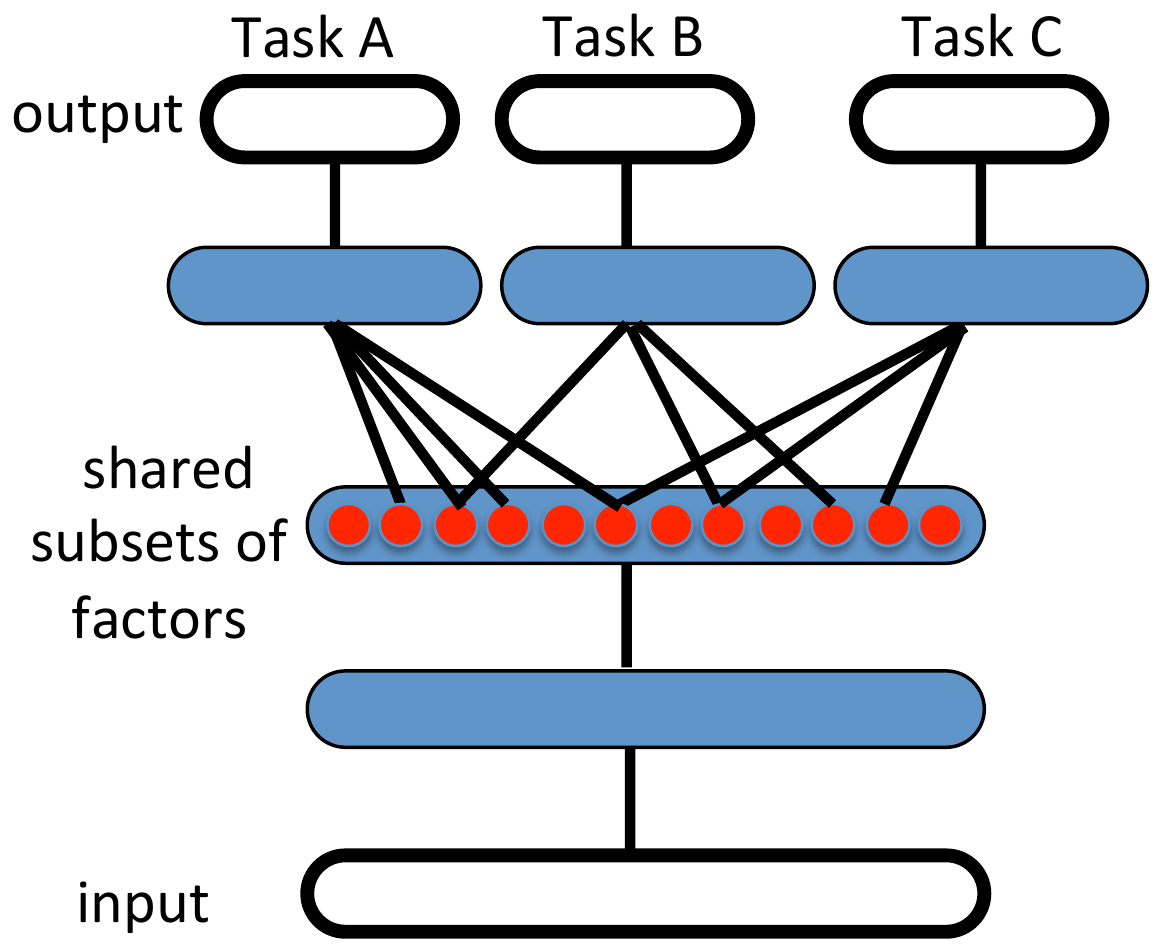}}
\vsC
\caption{\small Illustration of representation-learning
discovering explanatory
factors (middle hidden layer, in red), some explaining the
input (semi-supervised setting), and some explaining
target for each task.
Because these subsets overlap, sharing of statistical strength
helps generalization.} \label{fig:multi-task}.
\vsC
\end{figure}

Transfer learning is the ability of a learning algorithm to exploit
commonalities between different learning tasks in order to share statistical
strength, and {\em transfer knowledge} across tasks. As discussed below,
we hypothesize that representation learning algorithms have an advantage 
for such tasks because they
learn representations that capture underlying factors, a subset of which
may be relevant for each particular task, as illustrated
in Figure~\ref{fig:multi-task}. This hypothesis seems confirmed
by a number of empirical results showing the strengths of representation
learning algorithms in transfer learning scenarios.

Most impressive are the two transfer learning challenges held in 2011 and
won by representation learning algorithms. First, the Transfer Learning
Challenge, presented at an ICML 2011 workshop of the same name, was won
using unsupervised 
layer-wise pre-training~\citep{UTLC+DL+tutorial-2011-small,UTLC+LISA-2011-small}. A
second Transfer Learning Challenge was held the same year and won
by~\citet{Goodfellow+all-NIPS2011}. Results were presented at NIPS 2011's
Challenges in Learning Hierarchical Models Workshop. 
\iffalse
See Section~\ref{sec:transfer}
for a longer discussion and more pointers to other related results showing off
the natural ability of representation learning algorithms to generalize
to new classes, tasks, and domains.
\else
In the related {\em domain adaptation} setup, the target remains the same but the
input distribution changes~\citep{Glorot+al-ICML-2011-small,Chen-icml2012}. 
In the {\em multi-task learning} setup, representation learning has also been found
advantageous~\citet{Krizhevsky-2012-small,collobert:2011b}, because of shared factors
across tasks.


\vsD
\section{What makes a representation good?}
\label{sec:whatmakesitgood}
\vsA

\subsection{Priors for Representation Learning in AI}
\label{sec:priors}
\vsA

In~\citet{Bengio+Lecun-chapter2007}, one of us introduced the notion of AI-tasks,
which are challenging for current machine learning algorithms, and involve
complex but highly structured dependencies.
One reason why explicitly dealing with representations is interesting
is because they can be convenient to express many general priors about
the world around us, i.e., priors that are not task-specific but would
be likely to be useful for a learning machine to solve AI-tasks.
Examples of such general-purpose priors are the following:\\
$\bullet$ {\bf Smoothness}: assumes the function to be learned $f$ is s.t. $x\approx y$ generally implies $f(x)\approx f(y)$. This most basic prior is present in most machine learning, but is insufficient to get around the curse of dimensionality, see Section \ref{sec:smoothness}.\\
$\bullet$ {\bf Multiple explanatory factors}: the data generating distribution is generated by
different underlying factors, and for the most part what one learns about one
factor generalizes in many configurations of the other factors. The objective to
recover or at least disentangle these underlying factors of variation
is discussed in Section~\ref{sec:disentangling}. This assumption is behind the
idea of {\bf distributed representations}, discussed in Section \ref{sec:distributed} below.\\
$\bullet$ {\bf A hierarchical organization of explanatory factors}: the concepts that are useful for describing
the world around us can be defined in terms of other concepts, in a hierarchy,
with more {\bf abstract} concepts higher in the hierarchy,
defined in terms of less abstract ones.
This assumption is exploited with {\bf deep representations}, elaborated
in Section \ref{sec:depth} below.\\
$\bullet$ {\bf Semi-supervised learning}: 
  with inputs $X$ and target $Y$ to predict,
  a subset of the factors explaining $X$'s distribution explain much
  of $Y$, given $X$. Hence representations that are useful for $P(X)$
  tend to be useful when learning $P(Y|X)$, allowing sharing of statistical
  strength between the unsupervised and supervised learning tasks, see
  Section~\ref{sec:stacking}.\\
$\bullet$ {\bf Shared factors across tasks}: with many $Y$'s of
interest or many learning tasks in general, tasks (e.g., the corresponding $P(Y|X,{\rm task})$)
are explained by factors that are shared with other tasks, allowing sharing
of statistical strengths across tasks, as discussed in the previous
section (Multi-Task and Transfer Learning, Domain Adaptation).\\
$\bullet$ {\bf Manifolds}: probability mass concentrates near regions that have
a much smaller dimensionality than the original space where the data lives.
This is explicitly exploited in some of the auto-encoder algorithms and
other manifold-inspired algorithms described respectively in Sections 
\ref{sec:ae} and \ref{sec:manifold}.\\
$\bullet$ {\bf Natural clustering}: different values of
categorical variables such as object classes are associated with separate
manifolds. More precisely, the local variations on the manifold tend to
preserve the value of a category, and a linear interpolation between
examples of different classes in general involves going through a low density
region, i.e., $P(X|Y=i)$ for different $i$ tend to be well separated
and not overlap much. For example, this is exploited in the Manifold
Tangent Classifier discussed in Section~\ref{sec:leveraging-manifold}.
This hypothesis is consistent with the idea that humans have {\em named}
categories and classes because of such statistical structure (discovered
by their brain and propagated by their culture), and 
machine learning tasks often involves predicting such categorical variables.\\
$\bullet$ {\bf Temporal and spatial coherence}: 
consecutive (from a sequence) or spatially
nearby observations tend to be
associated with the same value of relevant categorical concepts, or
result in a small move on the surface of the high-density manifold. More generally,
different factors change at different temporal and spatial scales, and many categorical
concepts of interest change slowly. When attempting to capture such categorical
variables, this prior can be enforced by making the associated representations
slowly changing, i.e., penalizing changes in values over time or space. This prior
was introduced in~\cite{Becker92} and is discussed in Section~\ref{sec:slowness}.\\
$\bullet$
{\bf Sparsity}: for any given observation $x$, only a small fraction
  of the possible factors are relevant. In terms of representation, this
  could be represented by features that are often zero (as initially
  proposed by~\citet{Olshausen+Field-1996}), or by the fact that most of
  the extracted features are {\em insensitive} to small variations of $x$.
  This can be achieved with certain forms of priors on latent variables (peaked at 0), 
  or by using a non-linearity whose value is often
  flat at 0 (i.e., 0 and with a 0 derivative), or simply by penalizing the
  magnitude of the Jacobian matrix (of derivatives) of the function mapping
  input to representation. This is discussed in
  Sections~\ref{sec:sparse-coding} and~\ref{sec:ae}.\\
$\bullet$
{\bf Simplicity of Factor Dependencies}: in good high-level representations, the
factors are related to each other through simple, typically linear dependencies.
This can be seen in many laws of physics, and is assumed when plugging a linear
predictor on top of a learned representation.

We can view many of the above priors as ways to help the learner discover
and {\bf disentangle} some of the underlying (and a priori unknown) factors of variation 
that the data may reveal. This idea is pursued further in 
Sections~\ref{sec:disentangling}
and~\ref{sec:disentangling-algorithms}.

\vsD
\subsection{Smoothness and the Curse of Dimensionality}
\label{sec:smoothness}
\vsA

For AI-tasks, such as vision and NLP,
it seems hopeless to rely only on simple
parametric models (such as linear models) because they
cannot capture enough of the complexity of interest unless
provided with the appropriate feature space. Conversely,
machine learning researchers have sought flexibility in 
{\em local}\footnote{{\em local} in the sense that the value of the learned
function at $x$ depends mostly on training examples $x^{(t)}$'s close to $x$}
{\em non-parametric}
learners such as kernel machines with a fixed generic local-response kernel (such as
the Gaussian kernel).
Unfortunately, as argued at length by
\citet{Bengio+Monperrus-2005-short,Bengio-localfailure-NIPS-2006-small,Bengio+Lecun-chapter2007,Bengio-2009,Bengio-decision-trees10},
most of these algorithms only exploit the principle of {\em local
  generalization}, i.e., the assumption that the target function (to be
learned) is smooth enough, so they rely on examples to {\em explicitly map
  out the wrinkles of the target function}. Generalization is mostly achieved
by a form of local interpolation between neighboring training examples. 
Although smoothness can be a useful
assumption, it is insufficient to deal with the {\em curse of dimensionality},
because the number of such wrinkles (ups and downs of the target function)
may grow exponentially with the number of relevant interacting factors, when
the data are represented in raw input space.
We advocate learning algorithms that are flexible and 
non-parametric\footnote{We understand {\em non-parametric} as including all learning
algorithms whose capacity can be increased appropriately as the amount of 
data and its complexity demands it, e.g. including mixture models and neural networks
where the number of parameters is a data-selected hyper-parameter.}
but do not rely exclusively on the smoothness assumption. 
Instead, we propose to incorporate generic priors such as those enumerated
above into representation-learning algorithms. Smoothness-based learners
(such as kernel machines) and linear models can still be useful on top of
such learned representations. In fact,
the combination of learning a representation and kernel machine is equivalent
to {\em learning the kernel}, i.e., the feature space. Kernel
machines are useful, but they depend on a prior definition of
a suitable similarity metric, or a feature space in which naive similarity metrics suffice. We would like to use the data, along with very generic priors, to 
discover those features, or equivalently, a similarity function.

\vsD
\subsection{Distributed representations}
\label{sec:distributed}
\vsA

Good representations are {\em expressive}, 
meaning that a reasonably-sized learned representation can
capture a huge number of possible input configurations. A simple
counting argument helps us to assess the expressiveness of a model
producing a representation: how many parameters does it require compared to
the number of input regions (or configurations) it can distinguish?
Learners of one-hot representations, such as traditional clustering algorithms,
Gaussian mixtures, nearest-neighbor algorithms, decision trees,
or Gaussian SVMs all require $O(N)$ parameters (and/or $O(N)$ examples) to 
distinguish $O(N)$ input regions. One could naively believe that 
one cannot do better. However, RBMs,
sparse coding, auto-encoders or multi-layer neural networks can all represent up to $O(2^k)$
input regions using only $O(N)$ parameters (with $k$ the number of non-zero
elements in a sparse representation, and $k=N$ in non-sparse RBMs and other dense
representations). These are all {\em distributed} \footnote{
Distributed representations: where $k$ out of $N$
representation elements or feature values can be independently varied, e.g., they are not mutually exclusive.
Each concept is represented by having $k$ features being turned on or active, while
each feature is involved in representing many concepts.}
or sparse\footnote{Sparse representations: distributed representations where only
a few of the elements can be varied at a time, i.e., $k<N$.}  representations.
The generalization of clustering to distributed representations
is {\em multi-clustering}, where either several clusterings
take place in parallel or the same clustering is applied on different parts
of the input, such as in the very popular hierarchical feature extraction
for object recognition based on a histogram of cluster categories detected in different patches
of an image~\citep{Lazeb06,Coates2011b}. The exponential gain from distributed or sparse
representations is discussed further in section 3.2 (and Figure 3.2) of ~\citet{Bengio-2009}. It
comes about because each parameter (e.g. the parameters of one of the units in a sparse
code, or one of the units in a Restricted Boltzmann Machine) can be re-used
in many examples that are not simply near neighbors of each other, whereas
with local generalization, different regions in input space are basically associated with their
own private set of parameters, e.g., as in decision trees, nearest-neighbors, Gaussian SVMs, etc.
In a distributed representation, an exponentially
large number of possible {\em subsets} of features or hidden units can be activated in response 
to a given input. In a single-layer model, each feature 
is typically associated with a preferred input direction, corresponding to
a hyperplane in input space, and the {\em code} or representation 
associated with that input is precisely the
pattern of activation (which features respond to the input, and how much).
This is in contrast with a non-distributed
representation such as the one learned by most clustering algorithms, e.g., k-means,
in which the representation of a given input vector is a one-hot code
identifying which one of a small number of cluster centroids best represents the 
input~\footnote{As discussed in~\citep{Bengio-2009},
things are only slightly better when allowing continuous-valued
membership values, e.g., in ordinary mixture models (with separate parameters
for each mixture component), but the difference
in representational power is still exponential~\citep{Montufar+Morton-2012}.
The situation may also seem better with a decision tree, where each given input 
is associated with a one-hot code over the tree leaves, which deterministically
selects associated ancestors (the path from root to node). Unfortunately,
the number of different regions represented (equal to the number of leaves
of the tree) still only grows linearly with
the number of parameters used to specify it~\citep{Bengio+Delalleau-ALT-2011-short}.
}.

\vsD
\subsection{Depth and abstraction}
\label{sec:depth}
\vsA

Depth is a key aspect to representation learning strategies we consider in
this paper. As we will discuss, deep architectures are often challenging to
train effectively and this has been the subject of much recent research
and progress. However, despite these challenges, they carry two
significant advantages that motivate our long-term interest in discovering
successful training strategies for deep architectures. These advantages
are: (1) deep architectures promote the {\em re-use} of features, and (2)
deep architectures can potentially lead to progressively more {\em abstract}
features at higher layers of representations (more removed from the data).

{\bf Feature re-use.} The notion of re-use, which explains the power of distributed
representations, is also at the heart of the theoretical advantages
behind {\em deep learning}, i.e., constructing multiple levels of representation
or learning a hierarchy of features. The depth of a circuit is the length of the
longest path from an input node of the circuit to an output node of the circuit.
The crucial property of a deep circuit is that its number of paths, i.e., {\em ways to
re-use different parts}, can grow exponentially with its depth.
Formally, one can change the depth of a given circuit by changing the definition of what each 
node can
compute, but only by a constant factor. The typical computations we allow in each
node include: weighted sum, product, artificial neuron model (such as a monotone
non-linearity on top of an affine transformation), computation of a kernel,
or logic gates.
Theoretical results clearly show families
of functions where a deep representation can be exponentially more efficient
than one that is insufficiently deep 
\citep{Hastad86-small,Hastad91,Bengio-localfailure-NIPS-2006-small,Bengio+Lecun-chapter2007,Bengio+Delalleau-ALT-2011-short}.
If the same family of functions can be represented with fewer parameters
(or more precisely with a smaller VC-dimension), learning theory
would suggest that it can be learned with fewer examples, yielding improvements in both 
{\em computational} efficiency (less nodes to visit)
and {\em statistical} efficiency (less parameters to learn, and
re-use of these parameters over many different kinds of inputs).

{\bf Abstraction and invariance.} 
Deep architectures can lead to abstract representations because more
abstract concepts can often be constructed in terms of less abstract
ones. In some cases, such as in the convolutional neural
network~\citep{LeCun98-small}, we build this abstraction in explicitly via a
pooling mechanism (see section~\ref{sec:convol}).  More abstract concepts
are generally {\em invariant} to most local changes of the input. That
makes the representations that capture these concepts generally highly
non-linear functions of the raw input.  This is obviously true of
categorical concepts, where more abstract representations detect categories
that cover more varied phenomena (e.g. larger manifolds with more wrinkles)
and thus they potentially have greater predictive power. Abstraction can
also appear in high-level continuous-valued attributes that are only
sensitive to some very specific types of changes in the input. Learning
these sorts of invariant features has been a long-standing goal in pattern
recognition.

\vsD
\subsection{Disentangling Factors of Variation}
\label{sec:disentangling}
\vsA

Beyond being \emph{distributed} and \emph{invariant}, we would like our
representations to {\em disentangle the factors of variation}. Different
explanatory factors of the data tend to change independently of each other
in the input distribution, and only a few at a time tend to change when one
considers a sequence of consecutive real-world inputs.

Complex data arise from the rich interaction of many sources. These factors
interact in a complex web that can complicate AI-related tasks such as
object classification. For example, an image is composed of the interaction
between one or more light sources, the object shapes and the material
properties of the various surfaces present in the image. Shadows from
objects in the scene can fall on each other in complex patterns, creating
the illusion of object boundaries where there are none and dramatically
effect the perceived object shape. How can we cope with these complex
interactions? How can we \emph{disentangle} the objects and their shadows?
Ultimately, we believe the approach we adopt for overcoming these
challenges must leverage the data itself, using vast quantities of
unlabeled examples, to learn representations that separate the various
explanatory sources.  Doing so should give rise to a representation
significantly more robust to the complex and richly structured variations
extant in natural data sources for AI-related tasks.

It is important to distinguish between the related but distinct goals of
learning invariant features and learning to disentangle explanatory
factors. The central difference is the preservation of
information. Invariant features, by definition, have reduced sensitivity in
the direction of invariance. This is the goal of building features that are
insensitive to variation in the data that are uninformative to the task at
hand. Unfortunately, it is often difficult to determine \emph{a priori}
which set of features and variations will ultimately be relevant to the task at
hand. Further, as is often the case in the context of deep learning methods,
the feature set being trained may be destined to be
used in multiple tasks that may have distinct subsets of relevant features.
Considerations such as these lead us to the conclusion that the most robust
approach to feature learning is {\em to disentangle as many factors as
  possible, discarding as little information about the data as is
  practical}. If some form of dimensionality reduction is desirable, then
we hypothesize that the local directions of variation least represented in
the training data should be first to be pruned out (as in PCA, for example,
which does it globally instead of around each example).

\vsD
\subsection{Good criteria for learning representations?}
\vsA

One of the challenges of representation learning that distinguishes it from
other machine learning tasks such as classification is the difficulty in
establishing a clear objective, or target for training. In the case of
classification, the objective is (at least conceptually) obvious, we want
to minimize the number of misclassifications on the training dataset. In
the case of representation learning, our objective is far-removed from the
ultimate objective, which is typically learning a classifier or some other
predictor. Our problem is reminiscent of the credit assignment problem
encountered in reinforcement learning. We have proposed that
a good representation is one that disentangles the underlying factors
of variation, but how do we translate that into appropriate training criteria?
Is it even necessary to do anything but maximize likelihood under a good model
or can we introduce priors such as those enumerated above
(possibly {\em data-dependent} ones) that help
the representation better do this disentangling? This question remains
clearly open but is discussed in more detail in Sections~\ref{sec:disentangling}
and~\ref{sec:disentangling-algorithms}.

\vsD
\section{Building Deep Representations \note{YB}}
\label{sec:stacking}
\vsA

In 2006, a breakthrough in feature learning and deep 
learning was initiated by Geoff Hinton and quickly followed up
in the same year~\citep{Hinton06,Bengio-nips-2006-small,ranzato-07-small},
and soon after by \citet{HonglakL2008-small} and many more later.
It has been
extensively reviewed and discussed in~\citet{Bengio-2009}.  A central idea,
referred to as {\em greedy layerwise unsupervised pre-training}, was to learn 
a hierarchy of features one level at a time, using unsupervised feature
learning to learn a new transformation at each level to be composed with the previously learned 
transformations; essentially, each iteration of unsupervised feature learning adds one layer of weights to a deep 
neural network. Finally, the set of layers could be combined to initialize
a deep supervised predictor, such as a neural network classifier, or a deep
generative model, such as a Deep Boltzmann
Machine~\citep{Salakhutdinov+Hinton-2009-small}.


This paper is mostly about feature learning algorithms that can be used
to form deep architectures. In particular, it was empirically observed 
that layerwise {\em stacking} of feature extraction often yielded better
representations, e.g., in terms of classification error~\citep{Larochelle-jmlr-toappear-2008,Erhan+al-2010},
quality
of the samples generated by a probabilistic model~\citep{Salakhutdinov+Hinton-2009-small}
or in terms of the {\em invariance} properties of the learned features~\citep{Goodfellow2009-short}. Whereas this section focuses on the idea of stacking single-layer models,
Section~\ref{sec:global} follows up with a discussion on joint training of
all the layers.

After greedy layerwise unsuperivsed pre-training, 
the resulting deep features can be used either as input to a standard supervised
machine learning predictor (such as an SVM) or as initialization for a deep supervised neural
network (e.g., by appending a logistic regression layer or purely supervised layers of
a multi-layer neural network). The layerwise procedure can also be applied
in a purely {\em supervised} setting, called the {\em greedy layerwise supervised 
pre-training}~\citep{Bengio-nips-2006-small}. For example, after the first one-hidden-layer
MLP is trained, its output layer is discarded and another one-hidden-layer MLP can
be stacked on top of it, etc. Although results reported in~\citet{Bengio-nips-2006-small}
were not as good as for unsupervised pre-training, they were nonetheless better
than without pre-training at all. Alternatively, the {\em outputs} of the previous
layer can be fed as {\em extra inputs} for the next layer (in addition to the raw input), 
as successfully
done in~\citet{Yu+al-2010}. Another variant~\citep{Seide-et-al-ASRU2011} pre-trains
in a supervised way all the previously added layers at each step of the iteration, and in their
experiments this discriminant variant yielded better results than unsupervised pre-training.

Whereas combining single layers into a supervised model is straightforward,
it is less clear how layers pre-trained by unsupervised learning should be combined to form
a better {\em unsupervised} model. We cover here some of the approaches to do so, but
no clear winner emerges and much work has to be done to validate existing proposals or
improve them.

The first proposal was to stack pre-trained RBMs into a 
Deep Belief Network~\citep{Hinton06} or DBN,
where the top layer is interpreted as an RBM and the lower layers as a directed sigmoid belief network.
However, it is not clear how to approximate maximum likelihood training to further optimize
this generative model. One option is the wake-sleep algorithm~\citep{Hinton06} but more work
should be done to assess the efficiency of this procedure in terms of improving the generative
model. 

The second approach that has been put forward is to combine the RBM parameters
into a Deep Boltzmann Machine (DBM), by basically halving the RBM weights to obtain the DBM
weights~\citep{Salakhutdinov+Hinton-2009-small}. The DBM can then be trained by approximate maximum likelihood as
discussed in more details later (Section~\ref{sec:DBM}). This joint training has
brought substantial improvements, both in terms of
likelihood and in terms of classification performance of the resulting deep feature 
learner~\citep{Salakhutdinov+Hinton-2009-small}. 

Another early approach was to stack RBMs or auto-encoders into a {\em deep
  auto-encoder}~\citep{Hinton-Science2006}.  If we have a series of
encoder-decoder pairs $(f^{(i)}(\cdot),g^{(i)}(\cdot))$, then the overall
encoder is the composition of the encoders, $f^{(N)}(\ldots
f^{(2)}(f^{(1)}(\cdot)))$, and the overall decoder is its ``transpose''
(often with transposed weight matrices as well), $g^{(1)}(g^{(2)}(\ldots
f^{(N)}(\cdot)))$. The deep auto-encoder (or its regularized version, as
discussed in Section~\ref{sec:ae}) can then be jointly trained, with all
the parameters optimized with respect to a global reconstruction error criterion. More
work on this avenue clearly needs to be done, and it was probably avoided
by fear of the challenges in training deep feedforward networks, discussed
in the Section~\ref{sec:global} along with very encouraging recent results.

Yet another recently proposed approach to training deep architectures
~\citep{Ngiam-ICML2011} is to
consider the iterative construction of a {\em free energy function} (i.e., with no explicit
latent variables, except possibly for a top-level layer of hidden units) for a deep architecture
as the composition of transformations associated with lower layers, followed
by top-level hidden units. The question is
then how to train a model defined by an arbitrary parametrized (free) energy function.
\citet{Ngiam-ICML2011} have used Hybrid Monte Carlo~\citep{Neal93b}, but other
options include contrastive divergence~\citep{Hinton99-small,Hinton06},
score matching~\citep{Hyvarinen-2005-small,HyvarinenA2008}, denoising score 
matching~\citep{Kingma+LeCun-2010-small,Vincent-NC-2011-small},
ratio-matching~\citep{Hyvarinen-2007}
and noise-contrastive estimation~\citep{Gutmann+Hyvarinen-2010-small}.

\vsD
\section{Single-layer learning modules}
\vsA

Within the community of researchers interested in representation learning,
there has developed two broad parallel lines of inquiry: one rooted in
probabilistic graphical models and one rooted in neural networks.
Fundamentally, the difference between these two paradigms is whether the
layered architecture of a deep learning model is to be interpreted as
describing a probabilistic graphical model or as describing a computation
graph. In short, are hidden units considered latent random variables or as
computational nodes?

To date, the dichotomy between these two paradigms has remained in the background, perhaps
because they appear to have more characteristics in common than separating them.
We suggest that this is likely a function of the fact that much recent progress in both
of these areas has focused on \emph{single-layer greedy learning modules} and the similarities
between the types of single-layer models that have been explored: mainly, the restricted
Boltzmann machine (RBM) on the probabilistic side, and the auto-encoder variants on the neural
network side. Indeed, as shown by one of us~\citep{Vincent-NC-2011-small} and others~\citep{Swersky-ICML2011},
in the case of the restricted Boltzmann machine, training the model via an inductive
principle known as score matching~\citep{Hyvarinen-2005-small} (to be discussed in sec. \ref{sec:pseudolike}) is essentially identical to applying a regularized reconstruction objective to an auto-encoder. Another 
strong link between pairs of models on both sides of this divide
is when the computational graph 
for computing representation in the neural network model
corresponds exactly to the computational graph 
that corresponds to inference in the probabilistic model, 
and this happens to also correspond to the structure
of graphical model itself (e.g., as in the RBM).

The connection between these two paradigms becomes more tenuous when we consider deeper models where,
in the case of a probabilistic model, exact inference typically becomes intractable. 
In the case of deep models, the computational graph diverges from the structure of the model. 
For example, in the case of a deep Boltzmann machine, unrolling variational (approximate) 
inference into a computational graph results in a recurrent graph structure. 
We have performed preliminary exploration~\citep{Savard-master-small} 
of deterministic variants of deep auto-encoders 
whose computational graph is similar to that of a deep Boltzmann machine (in fact very 
close to the mean-field variational approximations associated with the Boltzmann machine), 
and that is one interesting intermediate point to explore (between the deterministic 
approaches and the graphical model approaches).

In the next few sections we will review the major developments in
single-layer training modules used to support feature learning and
particularly deep learning. We divide these sections between (Section~\ref{sec:prob-models}) the
probabilistic models, with inference and training schemes that
directly parametrize the generative -- or \emph{decoding} -- pathway and (Section~\ref{sec:direct})
the typically neural network-based models that directly parametrize the
\emph{encoding} pathway. Interestingly, some models, like Predictive Sparse 
Decomposition (PSD) \citep{koray-psd-08-small} inherit both properties, and will also be 
discussed (Section~\ref{sec:PSD}). We then present a different view
of representation learning, based on the associated geometry and the
manifold assumption, in Section~\ref{sec:manifold}.

First, let us consider an unsupervised single-layer representation learning
algorithm spaning all three views: probabilistic, auto-encoder, and
manifold learning.

{\bf Principal Components Analysis} 

We will use probably the oldest feature extraction algorithm, principal
components analysis (PCA), to
illustrate the probabilistic, auto-encoder and manifold views of
representation-learning. PCA learns a linear transformation $h = f(x) =
W^Tx+b$ of input $x\in \R^{d_x}$, where the columns of $d_x \times d_h$
matrix $W$ form an orthogonal basis for the $d_h$ orthogonal directions of
greatest variance in the training data. 
The result is $d_h$ features (the components of representation $h$) that are
decorrelated. The three interpretations of PCA are the following:
a) it is related to \emph{probabilistic models}
(Section \ref{sec:prob-models}) such as probabilistic PCA, factor analysis
and the traditional multivariate Gaussian distribution
(the leading eigenvectors of the covariance matrix are the
principal components); b) the representation it learns is essentially the
same as that learned by a basic linear \emph{auto-encoder} (Section
\ref{sec:ae}); and c) it can be viewed as a simple linear form of
linear \emph{manifold learning} (Section \ref{sec:manifold}), i.e., characterizing
a lower-dimensional region in input space near which the data density is
peaked. Thus, PCA may be 
in the back of the reader's mind as a common thread relating these various
viewpoints.  
Unfortunately the expressive power of linear features is very limited: they
cannot be stacked to form deeper, more abstract representations since the
composition of linear operations yields another linear operation. Here, we
focus on 
recent algorithms that have been developed to extract \emph{non-linear}
features, which can be stacked in the construction of deep networks,
although some
authors simply insert a
non-linearity between learned single-layer linear
projections~\citep{Le-CVPR2011-small,Chen-icml2012}. 

Another rich family
of feature extraction techniques that this review does not cover in any
detail due to space constraints is Independent Component Analysis or ICA 
\citep{Jutten+Herault-91,BellSejnowski-97}. Instead, we refer 
the reader to~\citet{Hyvarinen-2001,hyvarinen-book2009}. Note that, while in the
simplest case (complete, noise-free) ICA yields linear features, in the
more general case it can be equated with a \emph{linear generative model}
with non-Gaussian independent latent variables, similar to sparse coding
(section~\ref{sec:sparse-coding}), which result
in \emph{non-linear features}. Therefore, ICA and its variants like
Independent
and Topographic ICA~\citep{Hyvarinen+al-01} can and have been used to build
deep networks~\citep{Le2010-short,Le-CVPR2011-small}: see
section~\ref{sec:convol}.  The notion of obtaining independent components
also appears similar to our stated goal of disentangling underlying
explanatory factors through deep networks. However, for complex real-world
distributions,
it is doubtful that the relationship between truly independent underlying factors and the observed high-dimensional data can be adequately characterized by a linear transformation.

\vsD
\section{Probabilistic Models \note{AC}}
\label{sec:prob-models}
\vsA

From the probabilistic modeling perspective, the question of feature
learning can be interpreted as an attempt to recover a parsimonious set of 
latent random variables that describe a distribution over the observed data.
We can express as $p(x,h)$ a probabilistic model over the joint space of the latent
variables, $h$, and observed data or visible variables $x$.
Feature values are conceived as the result of an inference process to
determine the probability distribution of the latent variables given the
data, i.e. $p(h \mid x)$, often referred to as the \emph{posterior} probability. Learning is conceived in term of
estimating a set of model parameters that (locally) maximizes the
regularized likelihood of the training data.
The probabilistic graphical model formalism gives us two possible modeling
paradigms in which we can consider the question of inferring latent
variables, directed and undirected graphical models, which differ
in their parametrization of the joint distribution $p(x,h)$, yielding
%
major impact on the nature and computational costs of both inference and
learning. 

\vsD
\subsection{Directed Graphical Models}
\label{sec:directed}
\vsA

{\em Directed latent factor models} separately parametrize the
conditional likelihood $p(x \mid h)$ and the {\em prior}
$p(h)$ to construct the joint distribution, $p(x,h)=p(x \mid h)p(h)$.
Examples of this decomposition include: Principal
Components Analysis (PCA)~\citep{Roweis-97-small,tipping99-small}, sparse
coding~\citep{Olshausen+Field-1996}, sigmoid belief networks~\citep{Neal92} 
and the newly introduced spike-and-slab sparse coding model~\citep{Goodfellow+all-NIPS2011}.

\vsD
\subsubsection{\bf Explaining Away}
\vsA

Directed models often leads to one important property: {\em explaining away}, i.e.,
\emph{a priori} independent causes of an event can become
non-independent given the observation of the event. Latent factor
models can generally be interpreted as latent \emph{cause} models,
where the $h$ activations cause the observed $x$. This
renders the \emph{a priori} independent $h$ to be
non-independent. As a consequence, recovering the posterior distribution of
$h$, $p(h \mid x)$ (which we use as a basis for feature representation), is 
often computationally challenging and can be entirely
intractable, especially when $h$ is discrete.

A classic example that illustrates the phenomenon is to imagine you are on
vacation away from home and you receive a phone call from the security
system company, telling you that the alarm
has been activated. You begin worrying your home has been burglarized, but
then you hear on the radio that a minor earthquake has been reported in the
area of your home. If you happen to know from prior experience that
earthquakes sometimes cause your home alarm system to activate, then
suddenly you relax, confident that your home has very likely not been
burglarized.

The example illustrates how the {\bf alarm activation} rendered
two otherwise entirely independent causes, {\bf burglarized} and
{\bf earthquake}, to become dependent -- in this case, the dependency is
one of mutual exclusivity. Since both {\bf burglarized} and
{\bf earthquake} are very rare events and both can cause {\bf alarm
  activation}, the observation of one {\em explains away} the other.
Despite the computational obstacles we
face when attempting to recover the posterior over $h$, explaining away
promises to provide a parsimonious $p(h \mid x)$, which can be an extremely
useful characteristic of a feature encoding scheme.
If one thinks of a representation as being
composed of various feature detectors and estimated attributes of the
observed input, it is useful to allow the different features to
compete and collaborate with each other to explain the input. This is
naturally achieved with directed graphical models, but can also 
be achieved with undirected models (see Section~\ref{sec:undirected})
such as Boltzmann machines 
if there are {\em lateral connections} between the corresponding units
or corresponding {\em interaction terms} in the energy function that
defines the probability model.

\vspace*{1mm}
{\bf Probabilistic Interpretation of PCA.}
PCA can be given a natural probabilistic interpretation~\citep{Roweis-97-small,tipping99-small} 
 as {\em factor analysis}:

\vsE
\small
\addtolength{\jot}{0.5mm}
\begin{eqnarray}
p(h) & = & \mathcal{N}(h; 0,\sigma_h^2 \mathbf{I}) \nonumber \\
p(x \mid h) & = & \mathcal{N}(x; W h+\mu_x, \sigma_x^2 \mathbf{I}),
\vsD
\end{eqnarray}
\addtolength{\jot}{-0.5mm}
\normalsize
\vsF

\noindent where $x \in \mathbb{R}^{d_x}$, $h \in
\mathbb{R}^{d_h}$, $\mathcal{N}(v; \mu,\Sigma)$ is the multivariate
normal density of $v$ with mean $\mu$ and covariance $\Sigma$, and
columns of $W$ span the same space as leading $d_h$ principal components, but are not
constrained to be orthonormal.

{\bf Sparse Coding. \note{AC YB}}
\label{sec:sparse-coding}
Like PCA, sparse coding has both a probabilistic and 
non-probabilistic interpretation. Sparse coding also
relates a latent representation $h$ (either a vector of random variables or
a feature vector, depending on the interpretation) to the data $x$ through
a linear mapping $W$, which we refer to as the {\em dictionary}. 
The difference between sparse coding and PCA is that
sparse coding includes a penalty to ensure a
{\em sparse} activation of $h$ is used to encode each input $x$.
From a non-probabilistic perspective, 
sparse coding can be seen as recovering the code or feature vector associated with a new
input $x$ via:

\vsD
\small
\begin{equation}
h^{*} = f(x) = \argmin_h \|x - W h\|^2_2 + \lambda \|h\|_1,
\label{eq:sparse-coding-map}
\vsC
\end{equation}
\normalsize
Learning the dictionary $W$ can be accomplished by optimizing the
following training criterion with respect to $W$:

\vsC
\small
\begin{equation}
 \mathcal{J}_{\mbox{\tiny SC}} = \sum_t \|x^{(t)} - W h^{*(t)}\|^2_2,
\label{eq:sparse-coding-cost}
\vsC
\end{equation}
\normalsize
where $x^{(t)}$ is the $t$-th example and $h^{*(t)}$ is the corresponding sparse code determined
by Eq.~\ref{eq:sparse-coding-map}. $W$ is usually
constrained to have unit-norm columns (because one can
arbitrarily exchange scaling of column $i$ with scaling of $h^{(t)}_i$,
such a constraint is necessary for the L1 penalty to have any effect). 

The probabilistic interpretation of sparse coding differs from that
of PCA, in that instead of a Gaussian prior
on the latent random variable $h$, we use a sparsity inducing Laplace prior
(corresponding to an L1 penalty):
\vsD
\small
\begin{eqnarray}
p(h) & = & \prod_i^{d_h} \frac{\lambda}{2} \exp (-\lambda |h_{i}|) \nonumber \\
p(x \mid h) & = & \mathcal{N}(x; W h+\mu_x, \sigma_x^2 \mathbf{I}).
\vsC
\end{eqnarray}
\vsG
\normalsize 

\noindent In the case of sparse coding, because we will seek a sparse
representation (i.e., one with many features set to exactly zero), we will
be interested in recovering the MAP (maximum \emph{a posteriori} value of
$h$: i.e. $h^{*} = \argmax_h p(h \mid x)$ rather than its expected value
$\EE{h|x}$. Under this interpretation, dictionary learning
proceeds as maximizing the likelihood of the data \emph{given these MAP
  values of $h^{*}$}: \mbox{$\argmax_{W} \prod_{t} p(x^{(t)}\mid h^{*(t)})$}
subject to the norm constraint on $W$. Note that this parameter learning
scheme, subject to the MAP values of the latent $h$, is not standard
practice in the probabilistic graphical model literature. Typically the
likelihood of the data \mbox{$p(x) = \sum_{h}p(x\mid h) p(h)$} is maximized
directly. In the presence of latent variables, expectation
maximization is employed where the parameters are
optimized with respect to the marginal likelihood, i.e., summing or
integrating the joint log-likelihood over the all values of the latent
variables under their posterior $P(h\mid x)$, rather than considering only
the single MAP value of $h$. The theoretical properties of this form of parameter
learning are not yet well understood but seem to work well in practice
(e.g. k-Means vs Gaussian mixture models and Viterbi training for HMMs).
Note also that the interpretation of sparse coding as a MAP estimation can
be questioned~\citep{Gribonval-2011}, because even though the
interpretation of the L1 penalty as a log-prior is a possible
interpretation, there can be other Bayesian interpretations compatible with
the training criterion.

Sparse coding is an excellent example of the power of explaining away. 
Even with a very overcomplete dictionary\footnote{Overcomplete: with more dimensions of $h$ than dimensions of $x$.},
the MAP inference process used in sparse coding to find
$h^{*}$ can pick out the most appropriate bases and zero the others,
despite them having a high degree of correlation with the input. This
property arises naturally in directed graphical models such as sparse coding and
is entirely owing to the explaining away effect. It is not seen in commonly used
undirected probabilistic models such as the RBM, nor
is it seen in parametric feature encoding methods such as 
auto-encoders. The trade-off is that, compared to methods such as RBMs and
auto-encoders, inference in sparse coding involves an extra inner-loop of
optimization to find $h^{*}$ with a corresponding increase in the
computational cost of feature extraction. Compared to auto-encoders and
RBMs, the code in sparse coding is a free variable for each example, and 
in that sense the implicit encoder is non-parametric.

One might expect that the parsimony of the sparse coding
representation and its explaining away effect would be advantageous and indeed it
seems to be the case. \citet{Coates2011b} demonstrated on the CIFAR-10
object classification task~\citep{KrizhevskyHinton2009-small} with a patch-base feature extraction pipeline,
that in the regime with few ($< 1000$) labeled training examples per
class, the sparse coding representation significantly outperformed other
highly competitive encoding schemes.
Possibly because of these properties, and because of the very computationally efficient
algorithms that have been proposed for it (in comparison with the general
case of inference in the presence of explaining away),
sparse coding enjoys considerable
popularity as a feature learning and encoding paradigm. There are numerous
examples of its successful application as a feature representation scheme,
including natural image
modeling~\citep{RainaR2007-small,koray-psd-08-small,Coates2011b,Yu+Lin+Lafferty-2011-short},
audio classification~\citep{Grosse-2007-small}, 
NLP~\citep{Bradley+Bagnell-2009-small}, as well as being a very successful
model of the early visual cortex~\citep{Olshausen+Field-1996}. Sparsity
criteria can also be generalized successfully to yield groups of 
features that prefer to all be zero, but if one or a few of them are active 
then the penalty for activating others in the group is small.
Different {\em group sparsity} patterns can incorporate different
forms of prior knowledge~\citep{Koray-08-small,Jenatton-2009,Bach2011,gregor-nips-11-small}.

{\bf Spike-and-Slab Sparse Coding.}  Spike-and-slab sparse coding (S3C) is
one example of a promising variation on sparse coding for feature
learning~\citep{Goodfellow2012}. The S3C model possesses a set of latent
binary {\em spike} variables together with a a set of latent real-valued
{\em slab} variables. The activation of the spike variables dictates the
sparsity pattern.  S3C has been applied to the CIFAR-10 and CIFAR-100
object classification tasks~\citep{KrizhevskyHinton2009-small}, and shows
the same pattern as sparse coding of superior performance in the regime of
relatively few ($< 1000$) labeled examples per
class~\citep{Goodfellow2012}. In fact, in both the CIFAR-100 dataset (with
500 examples per class) and the CIFAR-10 dataset (when the number of
examples is reduced to a similar range), the S3C representation actually
outperforms sparse coding representations. This advantage was revealed
clearly with S3C winning the NIPS'2011 Transfer Learning
Challenge~\citep{Goodfellow+all-NIPS2011}.

\vsD
\subsection{Undirected Graphical Models}
\label{sec:undirected}
\vsA

Undirected graphical models, also called Markov random fields (MRFs), parametrize
the joint $p(x,h)$ through a product of unnormalized 
non-negative \emph{clique potentials}: 

\vsC
\small
\begin{equation}
p(x,h) = \frac{1}{Z_\theta} \prod_i \psi_i(x) \prod_j \eta_j(h) \prod_k \nu_k(x,h)
\vsC
\end{equation}
\normalsize
where $\psi_i(x)$, $\eta_j(h)$ and $\nu_k(x,h)$ are the clique potentials
describing the interactions between the visible elements, between the
hidden variables, and those interaction between the visible and hidden
variables respectively. The partition function $Z_\theta$ ensures that the
distribution is normalized. Within the context of unsupervised feature
learning, we generally see a particular form of Markov random field called
a Boltzmann distribution with clique potentials constrained to be positive:

\vsD
\small
\begin{equation}
p(x,h) = \frac{1}{Z_\theta} \exp \left( -\energy_\theta(x,h) \right),
\label{eq:BM_p(x,h)}
\vsC
\end{equation}
\normalsize
where $\energy_\theta(x,h)$ is the energy function and contains the interactions
described by the MRF clique potentials and $\theta$ are the model
parameters that characterize these interactions.

The Boltzmann machine was originally defined as a network of
symmetrically-coupled binary random variables or units. These
stochastic units can be divided into two groups: (1) the \emph{visible}
units $x \in \{0,1\}^{d_x}$ that represent the data, and (2) the
\emph{hidden} or latent
units $h \in \{0,1\}^{d_h}$ that mediate dependencies between the visible units
through their mutual interactions. The pattern of interaction is specified
through the energy function:

\vsD
\small
\begin{equation}
\energy^{\mathrm{BM}}_\theta(x,h) = -\frac{1}{2}x^TUx-\frac{1}{2}h^TVh-x^TWh-b^Tx-d^Th,
\label{eq:boltzmann_energy}
\vsC
\end{equation}
\normalsize
\vsC

\noindent where $\theta = \{U,V,W,b,d\}$ are the model parameters which respectively
encode the visible-to-visible interactions, the hidden-to-hidden
interactions, the visible-to-hidden interactions, the visible
self-connections, and the hidden self-connections (called biases). To
avoid over-parametrization, the diagonals of $U$ and $V$ are set to zero.

The Boltzmann machine energy function
specifies the probability distribution over $[x,h]$, via
the Boltzmann distribution, Eq.~\ref{eq:BM_p(x,h)}, with the partition
function $Z_\theta$ given by:

\vsD
\small
\begin{equation}
Z_\theta = \sum_{x_1 = 0}^{x_1 = 1}\cdots \sum_{x_{d_x} = 0}^{x_{d_x} = 1}
\sum_{h_1 = 0}^{h_1 = 1}\cdots \sum_{h_{d_h} = 0}^{h_{d_h} = 1} \exp\left(-\energy_\theta^{\mathrm{BM}}(x,h;\theta)\right).
\vsC
\end{equation}
\normalsize
This joint probability distribution gives rise to the set of conditional
distributions of the form:

\vsD
\small
\begin{align}
P(h_i \mid x, h_{\setminus i}) & = 
    \sigmoid
    \left( 
        \sum_{j} W_{ji}x_{j} + \sum_{i' \neq i} V_{ii'} h_{i'} + d_{i} 
    \right) \\ 
P(x_j \mid h, x_{\setminus j}) & = 
    \sigmoid
    \left( 
        \sum_{i}W_{ji}x_{j} + \sum_{j' \neq j} U_{jj'} x_{j'} + b_{j} 
    \right).
\vsC
\end{align}
\normalsize
In general, inference in the Boltzmann machine is intractable. For example,
computing the conditional probability of $h_i$ given the visibles, $P(h_i
\mid x)$, requires marginalizing over the rest of the hiddens, which implies
evaluating a sum with $2^{d_h-1}$ terms:

\vsD
\small
\begin{equation}
P(h_i \mid x) = \sum_{h_1 = 0}^{h_1 = 1}\cdots \sum_{h_{i-1} = 0}^{h_{i-1} = 1}
\sum_{h_{i+1} = 0}^{h_{i+1} = 1} \cdots \sum_{h_{d_h} = 0}^{h_{d_h} = 1} P(h \mid x)
\vsC
\end{equation}
\normalsize
However with some judicious choices
in the pattern of interactions between the visible and hidden units, more
tractable subsets of the model family are possible, as we discuss next.
 
{\bf Restricted Boltzmann Machines (RBMs).}
\label{sec:RBM}
The RBM is likely the most popular subclass of
Boltzmann machine~\citep{Smolensky86}. It is defined by restricting the interactions in the
Boltzmann energy function, in Eq. \ref{eq:boltzmann_energy}, to only those
between $h$ and $x$, i.e. $\energy_\theta^{\mathrm{RBM}}$ is $\energy_\theta^{\mathrm{BM}}$ 
with $U=\mathbf{0}$ and $V=\mathbf{0}$. As such, the
RBM can be said to form a \emph{bipartite} graph with the visibles and the
hiddens forming two layers of vertices in the graph (and no connection between
units of the same layer). With this
restriction, the RBM possesses the useful property that the conditional
distribution over the hidden units factorizes given the visibles:

\small
\vsD
\addtolength{\jot}{0.5mm}
\begin{eqnarray}
P(h \mid x) =& \prod_{i} P(h_{i} \mid x) \nonumber \\
P(h_i=1\mid x) =&\; \sigmoid\left(\sum_{j} W_{ji} x_{j}+d_{i}\right).
\label{eq:RBM_p(h|x)}
\vsD
\end{eqnarray}
\addtolength{\jot}{-0.5mm}
\normalsize
Likewise, the conditional distribution over the visible units given the
hiddens also factorizes:

\small
\vsD
\addtolength{\jot}{-2.5mm}
\begin{align}
P(x \mid h) =& \prod_{j} P(x_{j} \mid h) \nonumber \\
P(x_j=1\mid h) =&\; \sigmoid\left(\sum_{i} W_{ji} h_{i}+b_{j}\right).
\vsD
\vspace*{-1.5mm}
\end{align}
\addtolength{\jot}{2.5mm}
\normalsize
This makes inferences readily tractable in RBMs. For example, the
RBM feature representation is taken to be the set of posterior marginals
$P(h_i \mid x)$, which, given the conditional independence described in
Eq. \ref{eq:RBM_p(h|x)}, are immediately available. Note that this is in
stark contrast to the situation with popular directed graphical models for
unsupervised feature extraction, where computing the posterior probability
is intractable. 

Importantly, the tractability of the RBM does not extend to its partition
function, which still involves summing an exponential number of terms.
It does imply however that we can limit the number of terms to
$\min\{2^{d_x},2^{d_h}\}$. Usually this is still an unmanageable number of terms
and therefore we must resort to approximate methods to deal with its
estimation.

It is difficult to overstate the impact the RBM has had to the fields of
unsupervised feature learning and deep learning. It has been used in a
truly impressive variety of applications, including fMRI image
classification \citep{SchmahT2009-small}, motion and spatial
transformations~\citep{TaylorHintonICML2009-small,Memisevic+Hinton-2010-small},
collaborative filtering \citep{SalakhutdinovR2007b-small} and natural image
modeling \citep{Ranzato2010b-short,Courville+al-2011-small}. 

\vsD
\subsection{Generalizations of the RBM to Real-valued data}
\vsA

Important progress has been made in the last few years in defining
generalizations of the RBM that better capture real-valued data,
in particular real-valued image data, by better modeling the
conditional covariance of the input pixels.
The standard RBM, as discussed above, is defined with both binary visible
variables $v \in \{0,1\}$ and binary latent variables $h \in \{0,1\}$. The
tractability of inference and learning in the RBM has inspired many authors
to extend it, via modifications of its energy function, to model other
kinds of data distributions.  In particular, there has been multiple
attempts to develop RBM-type models of real-valued data, where $x\in
\R^{d_x}$.  The most straightforward approach to modeling real-valued
observations within the RBM framework is the so-called \emph{Gaussian RBM}
(GRBM) where the only change in the RBM energy function is to the visible
units biases, by adding a bias term that is quadratic in the visible units
$x$. While it probably remains the most popular way to model real-valued
data within the RBM framework, \citet{Ranzato2010b-short} suggest that the
GRBM has proved to be a somewhat unsatisfactory model of natural
images. The trained features typically do not represent sharp edges that
occur at object boundaries and lead to latent representations that are not
particularly useful features for classification
tasks. \citet{Ranzato2010b-short} argue that the failure of the GRBM to
adequately capture the statistical structure of natural images stems from
the exclusive use of the model capacity to capture the conditional mean at
the expense of the conditional covariance. Natural images, they argue, are
chiefly characterized by the covariance of the pixel values, not by their
absolute values. This point is supported by the common use of preprocessing
methods that standardize the global scaling of the pixel values across
images in a dataset or across the pixel values within each image.

These kinds of concerns about the ability of the GRBM to model
natural image data has lead to the development of alternative RBM-based
models that each attempt to take on this objective of {\em better modeling
non-diagonal conditional covariances}. ~\citep{Ranzato2010b-short} introduced the
\emph{mean and covariance RBM} (mcRBM). Like the GRBM, the mcRBM is
a 2-layer Boltzmann machine that explicitly models the visible units as
Gaussian distributed quantities. However unlike the GRBM, the mcRBM
uses its hidden layer to independently parametrize both the mean and
covariance of the data through two sets of hidden units. The mcRBM is a
combination of the covariance RBM (cRBM) \citep{ranzato2010factored-small},
that models the conditional covariance, with the GRBM that captures
the conditional mean. While the GRBM has shown considerable
potential as the basis of a highly successful phoneme recognition system
\citep{dahl2010phonerec-small}, it seems that due to difficulties in
training the mcRBM, the model has been largely superseded by the mPoT
model. The mPoT model (\emph{mean-product of Student's T-distributions
  model}) ~\citep{ranzato+mnih+hinton:2010-short} is a combination of the
GRBM and the product of Student's T-distributions
model~\citep{Welling2003a-small}. It is an energy-based model where the
conditional distribution over the visible units conditioned on the hidden
variables is a multivariate Gaussian (non-diagonal covariance) and the
complementary conditional distribution over the hidden variables given the
visibles are a set of independent Gamma distributions.

The PoT model has recently been
generalized to the mPoT model~\citep{ranzato+mnih+hinton:2010-short} to include
nonzero Gaussian means by the addition of GRBM-like hidden units,
similarly to how the mcRBM generalizes the cRBM.
The mPoT model has been used to synthesize large-scale natural images
\citep{ranzato+mnih+hinton:2010-short} that show large-scale features and
shadowing structure. It has been used to model natural
textures~\citep{Kivinen2012-short} in a {\em tiled-convolution} configuration
(see section~\ref{sec:convol}).

Another recently introduced RBM-based model with the objective of having
the hidden units encode both the mean and covariance information is the
\emph{spike-and-slab} Restricted Boltzmann Machine
(ssRBM)~\citep{CourvilleAISTATS11-small,Courville+al-2011-small}. The ssRBM is
defined as having both a real-valued ``slab'' variable and a binary
``spike'' variable associated with each unit in the hidden layer.
The ssRBM has been demonstrated as a feature learning and extraction scheme
in the context of CIFAR-10 object classification~\citep{KrizhevskyHinton2009-small} from natural images and has
performed well in the
role~\citep{CourvilleAISTATS11-small,Courville+al-2011-small}. When trained
{\em convolutionally} (see Section~\ref{sec:convol}) on full CIFAR-10 natural images, the model demonstrated the
ability to generate natural image
samples that seem to capture the broad statistical structure of natural images
better than previous parametric generative models,
as illustrated with the samples of Figure~\ref{fig:ssRBM-samples}.

\begin{figure}[ht]
\vsC
    \centering
        \includegraphics[scale=0.45]{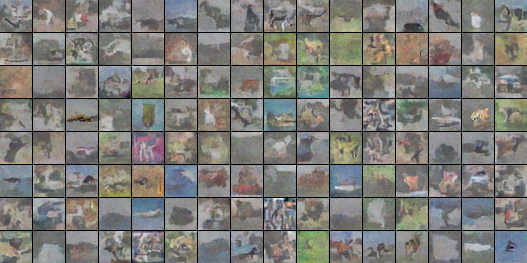}
\vspace*{1mm}

        \includegraphics[scale=0.45]{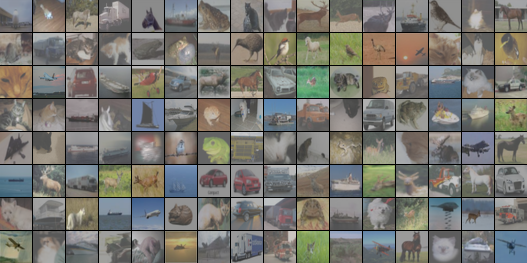}
\vsC
    \caption{ \small
    (Top) Samples from convolutionally trained \mbox{$\mu$-ssRBM} from~\citet{Courville+al-2011-small}.
    (Bottom) Images in CIFAR-10
    training set closest (L2 distance with contrast normalized training images) to {\em corresponding}
    model samples on top. The model does not appear to be
    overfitting particular training examples.
    }
    \label{fig:ssRBM-samples}
\vsE
\end{figure}

The mcRBM, mPoT and ssRBM each set out to model real-valued data such that
the hidden units encode not only the conditional mean of the data but also
its conditional covariance. Other than differences in the training schemes,
the most significant difference between these models is how they
encode their conditional covariance. While the mcRBM and the mPoT use the
activation of the hidden units to enforce constraints on the
covariance of $x$, 
the ssRBM uses the hidden unit to pinch the precision matrix along
the direction specified by the corresponding weight vector.
These two ways of modeling conditional covariance diverge when the
dimensionality of the hidden layer is significantly different from that of the input.
In the overcomplete setting, sparse activation with the ssRBM
parametrization permits variance only in the select directions of the sparsely
activated hidden units. This is a property the ssRBM shares with sparse coding
models~\citep{Olshausen+Field-1996,Grosse-2007-small}.
On the other hand, in the case of the mPoT or mcRBM, an overcomplete set of constraints on
the covariance implies that capturing arbitrary covariance along
a particular direction of the input requires decreasing potentially all
constraints with positive projection in that direction. This perspective would suggest that the mPoT and
mcRBM do not appear to be well suited to provide a sparse representation in
the overcomplete setting.


\vsD
\subsection{RBM parameter estimation}
\vsA

Many of the RBM training methods we discuss here are 
applicable to more general undirected graphical models, but are
particularly practical in the RBM setting. \citet{Freund+Haussler-94}
proposed a learning algorithm for harmoniums (RBMs) based on
projection pursuit. Contrastive 
Divergence~\citep{Hinton99-small,Hinton06} has been used most often to train RBMs,
and many recent papers use Stochastic Maximum Likelihood~\citep{Younes1999,Tieleman08-small}.

As discussed in Sec.~\ref{sec:directed}, in training probabilistic models parameters are typically adapted
in order to maximize the \emph{likelihood of the training data} (or
equivalently the log-likelihood, or its penalized version, which adds a regularization term). With $T$
training examples, the log likelihood is given by:

\vsF
\small
\begin{equation}
\sum_{t=1}^T \log P(x^{(t)}; \theta) = \sum_{t=1}^T \log \sum_{h \in \{0,1\}^{d_h}}P(x^{(t)}, h ; \theta).
\vsD
\end{equation}
\normalsize
\vsB

\noindent 
Gradient-based optimization requires its gradient, which for Boltzmann machines, 
is given by:

\vsF
\small
\begin{eqnarray}
\frac{\partial}{\partial\theta_{i}}\sum_{t=1}^{T}\log p(x^{(t)})
 &=& - \sum_{t=1}^{T}\E{p(h\mid x^{(t)})}{
     \frac{\partial}{\partial\theta_{i}}\energy_\theta^{\mathrm{BM}}(x^{(t)},h)} \nonumber \\
  &&+ 
     \sum_{t=1}^{T}\E{p(x,h)}{
     \frac{\partial}{\partial\theta_{i}}\energy_\theta^{\mathrm{BM}}(x,h)},
\label{eq:BM_llgrad}
\vsE
\end{eqnarray}
\normalsize
where we have the expectations with respect to $p(h^{(t)}\mid x^{(t)})$
in the {}``clamped'' condition (also called the positive phase), and over the full joint $p(x,h)$
in the {}``unclamped'' condition (also called the negative
phase). Intuitively, the gradient acts to locally move the model
distribution (the negative phase distribution) toward the data distribution
(positive phase distribution), by pushing down the energy of $(h,x^{(t)})$
pairs (for $h \sim P(h|x^{(t)})$) while pushing up the energy of $(h,x)$
pairs (for $(h,x) \sim P(h,x)$) until the two forces are in equilibrium,
at which point the sufficient statistics (gradient of the energy function)
have equal expectations with $x$ sampled from the training distribution or with 
$x$ sampled from the model.

The RBM conditional independence properties imply that the expectation in
the positive phase of
Eq.~\ref{eq:BM_llgrad} is tractable. The negative phase term -- arising from the
partition function's contribution to the log-likelihood gradient -- is more
problematic because the computation of the expectation over the joint is not
tractable. The various ways of dealing with the partition function's contribution to
the gradient have brought about a number of different training algorithms,
many trying to approximate the log-likelihood gradient. 

To approximate the expectation of the joint distribution
in the negative phase contribution to the gradient, it is natural to again consider
exploiting the conditional independence of the RBM in order to specify a Monte Carlo
approximation of the expectation over the joint:

\vsF
\small
\begin{equation}
\E{p(x,h)}{
     \frac{\partial}{\partial\theta_{i}}\energy_\theta^{\mathrm{RBM}}(x,h)}
   \approx
   \frac{1}{L}\sum_{l=1}^{L}\frac{\partial}{\partial\theta_{i}}\energy_\theta^{\mathrm{RBM}}(\tilde{x}^{(l)},\tilde{h}^{(l)}),
\label{eq:mc_dZ}
\vsC
\end{equation}
\normalsize 
\vsB

\noindent 
with the samples $(\tilde{x}^{(l)},\tilde{h}^{(l)})$ drawn by a
block Gibbs MCMC (Markov chain Monte Carlo) sampling procedure:

\vsF
\small
\addtolength{\jot}{0.5mm}
\begin{eqnarray}
\tilde{x}^{(l)} & \sim & P(x \mid \tilde{h}^{(l-1)}) \nonumber \\ 
\tilde{h}^{(l)} & \sim & P(h \mid \tilde{x}^{(l)}). \nonumber 
\vsE
\end{eqnarray}
\addtolength{\jot}{-0.5mm}
\normalsize
\vsF

\noindent Naively, for each gradient update step, one would start a Gibbs sampling
chain, wait until the chain converges to the equilibrium distribution and
then draw a sufficient number of samples to approximate the expected gradient with
respect to the model (joint) distribution in Eq. \ref{eq:mc_dZ}. Then
restart the process for the next step of approximate gradient ascent on the
log-likelihood. This procedure has the obvious flaw that waiting for the
Gibbs chain to ``burn-in'' and reach equilibrium anew for each gradient
update cannot form the basis of a practical training algorithm. Contrastive
Divergence~\citep{Hinton99-small,Hinton06}, Stochastic Maximum Likelihood~\citep{Younes1999,Tieleman08-small} 
and fast-weights persistent
contrastive divergence or FPCD~\citep{TielemanT2009-small} are all ways
to avoid or reduce the need for burn-in.

\vsC
\subsubsection{\bf Contrastive Divergence}
\vsA

Contrastive divergence (CD) estimation~\citep{Hinton99-small,Hinton06} 
estimates the negative phase expectation (Eq.~\ref{eq:BM_llgrad})
with a very short Gibbs chain (often just one
step) initialized \emph{at the training data used in the positive
  phase}. This reduces the variance of the gradient estimator and still
moves in a direction that pulls the negative chain samples towards
the associated positive chain samples.
Much has been written about the properties and alternative interpretations
of CD and its similarity to auto-encoder training, 
e.g. \citet{Perpinan+Hinton-2005-small,Yuille2005-small,Bengio+Delalleau-2009,sutskever2010convergence-small}.

\vsC
\subsubsection{\bf Stochastic Maximum Likelihood}
\vsA

The Stochastic Maximum Likelihood (SML) algorithm (also known as
persistent contrastive divergence or PCD)~\citep{Younes1999,Tieleman08-small} is
an alternative way to sidestep an extended burn-in of the negative phase
Gibbs sampler. At each gradient update, rather than initializing the Gibbs
chain at the positive phase sample as in CD, SML initializes the chain at
the last state of the chain used for the previous update. In other words, SML uses a
continually running Gibbs chain (or often a number of Gibbs chains run in parallel) from
which samples are drawn to estimate the negative phase expectation. Despite the
model parameters changing between updates, these changes should be small
enough that only a few steps of Gibbs (in practice, often one step is
used) are required to maintain samples from the equilibrium distribution of
the Gibbs chain, i.e. the model distribution.

A troublesome aspect of SML is that
it relies on the Gibbs chain to mix well (especially between modes) for
learning to succeed. Typically, as learning progresses and the weights of the
RBM grow, the ergodicity of the Gibbs sample begins to break down\footnote{When
weights become large, the estimated distribution is more peaky, and the
chain takes very long time to mix, to move from mode to mode, so that practically
the gradient estimator can be very poor. This is a serious chicken-and-egg problem
because if sampling is not effective, nor is the training procedure, which may
seem to stall, and yields even larger weights.}. If the
learning rate $\epsilon$ associated with gradient ascent 
$\theta \leftarrow \theta + \epsilon \hat{g}$
(with $E[\hat{g}]\approx\frac{\partial \log p_\theta(x)}{\partial \theta}$)
is not reduced to compensate,
then the Gibbs sampler will diverge from the model distribution and
learning will fail.
\citet{Desjardins+al-2010-small,Cho10IJCNN-small,Salakhutdinov-2010-small,Salakhutdinov-ICML2010-small} have all
considered various forms of tempered transitions to 
address the
failure of Gibbs chain mixing, and convincing solutions 
have not yet been clearly demonstrated. A recently introduced promising avenue
relies on depth itself, showing that mixing between modes is much easier
on deeper layers~\citep{Bengio-et-al-ICML2013} (Sec.\ref{sec:sampling-challenge}).

\citet{TielemanT2009-small} have proposed
quite a different approach to addressing potential mixing problems
of SML with their fast-weights persistent contrastive divergence
(FPCD), and it has also been exploited to train Deep Boltzmann Machines~\citep{Salakhutdinov-ICML2010-small}
and construct a pure sampling algorithm for RBMs~\citep{Breuleux+Bengio-2011}.
FPCD builds on the surprising but robust tendency of Gibbs
chains to mix better during SML learning than when the model parameters are
fixed. The phenomenon is rooted in the form of the likelihood gradient
itself (Eq.~\ref{eq:BM_llgrad}). The samples drawn from the SML Gibbs chain
are used in the negative phase of the gradient, which implies that the learning
update will slightly increase the energy (decrease the probability) of
those samples, making the region in the neighborhood of those samples less
likely to be resampled and therefore making it more likely that the samples
will move somewhere else (typically going near another mode).
Rather than drawing samples from the distribution of the current model (with parameters
$\theta$), FPCD exaggerates
this effect by drawing samples from a local perturbation of the model with
parameters $\theta^{*}$ and an update

\vsG
\small
\begin{equation}
\theta^{*}_{t+1} = (1-\eta) \theta_{t+1} + \eta \theta^{*}_t + \epsilon^{*} \frac{\partial}{\partial\theta_{i}}\left(\sum_{t=1}^{T}\log p(x^{(t)})\right),
\vsC
\end{equation}
\normalsize
where $\epsilon^{*}$ is the relatively large fast-weight learning rate
($\epsilon^{*} > \epsilon$) and $0 < \eta < 1$ (but near 1) is a forgetting factor
that keeps the perturbed model close to the current model. Unlike
tempering, FPCD does not converge to the model distribution as $\epsilon$
and $\epsilon^*$ go to 0, and further work is necessary to characterize the nature of
its approximation to the model distribution. Nevertheless, FPCD is a
popular and apparently effective means of drawing approximate samples from
the model distribution that faithfully represent its diversity, at the price
of sometimes generating spurious samples {\em in between two modes} (because
the fast weights roughly correspond to a smoothed view of the current model's
energy function). It
has been applied in a variety of applications~\citep{TielemanT2009-small,Ranzato2011-short,Kivinen2012-short}
and it has been transformed into a sampling algorithm~\citep{Breuleux+Bengio-2011}
that also shares this fast mixing property with {\em herding}~\citep{WellingUAI2009-small},
for the same reason, i.e., introducing {\em negative correlations} between
consecutive samples of the chain in order to promote faster mixing.

\vsC
\subsubsection{\bf Pseudolikelihood, Ratio-matching and More}
\vsA

\label{sec:pseudolike}
While CD, SML and FPCD are by far the most popular methods for training
RBMs and RBM-based models, all of these methods are perhaps most naturally
described as offering different approximations to maximum likelihood
training. There exist other inductive principles that are alternatives to
maximum likelihood that can also be used to train RBMs. In particular,
these include pseudo-likelihood~\citep{Besag75pseudolikelihood} and
ratio-matching~\citep{Hyvarinen-2007}. Both of these inductive principles
attempt to avoid explicitly dealing with the partition
function, and their asymptotic efficiency has been analyzed~\citep{Marlin11-small}. 
Pseudo-likelihood seeks to maximize the product of all
one-dimensional conditional distributions of the form $P(x_d|x_{\setminus d})$,
while ratio-matching can be interpreted as an extension of score matching~\citep{Hyvarinen-2005-small} to
discrete data types. Both methods amount to weighted
differences of the gradient of the RBM free energy\footnote{The
  free energy $\mathcal{F}(x;\theta)$ is the energy associated with the data marginal probability,
  $\mathcal{F}(x;\theta) = -\log P(x) -\log Z_{\theta}$ and is
  tractable for the RBM.} evaluated at a data point and at
neighboring points.
One potential drawback of these methods is that depending on the parametrization of the
energy function, their computational requirements may scale up to $O(n_d)$ worse than
 CD, SML, FPCD, or {\em denoising score 
matching}~\citep{Kingma+LeCun-2010-small,Vincent-NC-2011-small}, discussed below.
\citet{Marlin10Inductive-small} empirically compared all of these
methods (except denoising score matching)
on a range of classification, reconstruction and density modeling
tasks and found that, in general, SML provided the best combination of
overall performance and computational tractability. However, in a later
study, the same authors~\citep{Swersky-ICML2011} found denoising
score matching 
to be a competitive inductive principle both
in terms of classification performance (with respect to SML) and
in terms of computational efficiency (with respect to analytically
obtained score matching). Denoising score matching
is a special case of the denoising auto-encoder training
criterion (Section~\ref{sec:dae}) when the reconstruction
error residual equals a gradient, i.e., the score function associated
with an energy function, as shown in~\citep{Vincent-NC-2011-small}.

In the spirit of the Boltzmann machine gradient (Eq.~\ref{eq:BM_llgrad})
several approaches have been proposed to train energy-based
models. One is {\em noise-contrastive
  estimation}~\citep{Gutmann+Hyvarinen-2010-small}, in which the training
criterion is transformed into a {\em probabilistic classification problem}:
distinguish between (positive) training examples and (negative) noise
samples generated by a broad distribution (such as the Gaussian). Another
family of approaches, more in the spirit of Contrastive Divergence, relies
on distinguishing positive examples (of the training distribution) and
negative examples obtained by perturbations of the positive
examples~\citep{CollobertR2008-small,Antoine-al-2012-small,Weston+Bengio+Usunier-2010}. 

\vsD
\section{Directly Learning A Parametric Map from Input to Representation}
\label{sec:direct}
\vsA

Within the framework of probabilistic models adopted in
Section~\ref{sec:prob-models}, the learned representation is always associated with latent
variables, specifically with their posterior distribution given an observed
input $x$.  Unfortunately, this posterior distribution
tends to become very complicated and intractable if the model
has more than a couple of interconnected layers, whether in the directed or
undirected graphical model frameworks. It then becomes necessary to resort
to sampling or approximate inference techniques, and to pay the associated
computational and approximation error price. If the true posterior has a large
number of modes that matter then current inference techniques may face an unsurmountable
challenge or endure a potentially serious approximation. This is in addition to the
difficulties raised by the intractable partition function in undirected
graphical models.  Moreover a posterior \emph{distribution} over latent
variables is not yet a simple usable \emph{feature vector} that can for
example be fed to a classifier. So actual feature values are typically
\emph{derived} from that distribution, taking the latent variable's
expectation (as is typically done with RBMs), their marginal probability,
or finding their most likely value (as in sparse coding).  If we are to
extract stable deterministic numerical feature values in the end anyway, an
alternative (apparently) non-probabilistic feature learning paradigm that
focuses on carrying out this part of the computation, very efficiently, is
that of auto-encoders and other directly parametrized feature or
representation functions.  The commonality between these methods is that
they {\em learn a direct encoding, i.e., a parametric map from inputs to
  their representation}.

Regularized auto-encoders, discussed next, also involve learning 
a decoding function that maps back from representation to input space. 
Sections~\ref{sec:neighbors}
and~\ref{sec:slowness} discuss direct encoding methods that do not require
a decoder, such as semi-supervised
embedding~\citep{WestonJ2008-small} and slow feature analysis~\citep{wiskott:2002}.

\vsD
\subsection{Auto-Encoders}
\vsB


In the auto-encoder framework \citep{Lecun-these87,Bourlard88,hinton1994amd-small}, 
one starts by explicitly defining a
feature-extracting function in a specific parametrized closed form. This
function, that we will denote $f_\theta$, is called the {\bf encoder} 
and will allow the straightforward and efficient computation of
a feature vector $h = f_\theta(x)$ from an input $x$. 
For each example $x^{(t)}$ from a data set $\{x^{(1)}, \ldots, x^{(T)} \}$, we define

\vsD
\small
\begin{equation}
 h^{(t)} = f_\theta(x^{(t)})  
\label{eq:codes}
\vsD
\end{equation}
\normalsize 
\vsE

\noindent where $h^{(t)}$ is the \emph{feature-vector} or
\emph{representation} or \emph{code} computed from $x^{(t)}$.  Another
closed form parametrized function $g_\theta$, called the {\bf decoder},
maps from feature space back into input space, producing a {\bf
  reconstruction} $r = g_\theta(h)$. Whereas probabilistic models are
defined from an explicit probability function and are trained to maximize
(often approximately) the data likelihood (or a proxy), auto-encoders are
parametrized through their encoder and decoder and are trained using a
different training principle.  The set of parameters $\theta$ of the
encoder and decoder are learned simultaneously on the task of
reconstructing as well as possible the original input, i.e. attempting to
incur the lowest possible {\bf reconstruction error} $L(x,r)$ -- a measure
of the discrepancy between $x$ and its reconstruction $r$ -- {\em over
training examples}. Good generalization means low reconstruction error
at test examples, while having high reconstruction error for most other $x$ configurations.
To capture the structure of the
data-generating distribution, it is therefore important that something in
the training criterion or the parametrization prevents the auto-encoder
from learning the identity function, which has zero reconstruction
error everywhere. This is achieved through various means in the different
forms of auto-encoders, as described below in more detail, and we call
these {\em regularized auto-encoders}. A particular form of regularization
consists in constraining the code to have a low dimension, and this is
what the classical auto-encoder or PCA do. 

In summary, basic auto-encoder training consists in finding a value of
parameter vector $\theta$
minimizing reconstruction error

\vsF
\small
\begin{eqnarray}
\mathcal{J}_{\mbox{\tiny AE}}(\theta) &=& \sum_t L(x^{(t)} ,g_\theta(f_\theta(x^{(t)})))
\vsD
\label{eq:reconstruction-error}
\end{eqnarray}
\normalsize
\vsE

\noindent where $x^{(t)}$ is a training example.
This minimization is usually carried out by stochastic gradient descent as in the
training of Multi-Layer-Perceptrons (MLPs).
Since auto-encoders were primarily developed as MLPs
predicting their input, the
most commonly used forms for the encoder and decoder are affine mappings,
optionally followed by a non-linearity: 

\vsF
\small
\begin{eqnarray}
f_\theta(x) &=& s_f(b + W x) \label{eq:AE} \\
g_\theta(h) &=& s_g(d + W' h) \label{eq:AEg}
\vsD
\end{eqnarray}
\normalsize
\vsG

\noindent where $s_f$ and $s_g$ are the encoder and decoder activation functions
(typically the element-wise sigmoid or hyperbolic tangent non-linearity, or the identity function if
staying linear). The set of parameters of such a model is $\theta = \{ W,
b, W', d \}$ where $b$ and $d$ are called encoder and decoder bias vectors,
and $W$ and $W'$ are the encoder and decoder weight matrices. 

The choice of $s_g$ and $L$ depends largely on the input domain range
and nature, and are usually chosen so that $L$ returns a negative log-likelihood
for the observed value of $x$.
A natural choice for an unbounded domain is a linear decoder with a squared
reconstruction error, i.e. $s_g(a) = a$ and $L(x,r) = \|x-r\|^2$. 
If inputs are bounded between $0$ and $1$ however, ensuring a
similarly-bounded reconstruction can be achieved by using $s_g =
\sigmoid$. In addition if the
inputs are of a binary nature, a binary cross-entropy loss\footnote{$L(x,r)
  = - \sum_{i=1}^{d_x} x_i \log(r_i) + (1-x_i) \log (1-r_i)$} is sometimes used.

If both encoder and decoder use a sigmoid non-linearity, then $f_\theta(x)$ 
and $g_\theta(h)$ have the
exact \emph{same form} as the conditionals $P(h \mid v)$ and $P(v \mid h)$
of binary RBMs
(see Section \ref{sec:RBM}). This similarity motivated an initial study~\citep{Bengio-nips-2006-small} of the
possibility of replacing RBMs with
auto-encoders as the basic pre-training strategy for building deep
networks, as well as the comparative analysis of auto-encoder reconstruction
error gradient and contrastive divergence updates~\citep{Bengio+Delalleau-2009}.

One notable difference in the parametrization is that 
RBMs use a single weight matrix, which follows
naturally from their energy function, whereas the auto-encoder framework allows for a
different matrix in the encoder and decoder. In practice however, \emph{weight-tying} in which one defines $W'=W^T$ 
may be (and is most often) used, rendering the parametrizations identical. 
The usual training procedures however differ greatly between the two
approaches. A practical advantage of training
auto-encoder variants is that {\em they define a simple tractable 
optimization objective that can be used to monitor progress}. 

In the case of a linear auto-encoder (linear encoder and decoder) with squared reconstruction error,
minimizing Eq.~\ref{eq:reconstruction-error} learns the same
\emph{subspace}\footnote{Contrary to traditional PCA loading factors, but
  similarly to the parameters learned by probabilistic PCA, the 
  weight vectors learned by a linear auto-encoder are not constrained to form an orthonormal basis, nor to
  have a meaningful ordering. They will however span the same subspace.} as
PCA. This is also true when using a sigmoid nonlinearity
in the encoder~\citep{Bourlard88}, but not
if the weights $W$ and $W'$ are tied ($W'=W^T$), because
$W$ cannot be forced into being small and $W'$ large to achieve a linear encoder.

Similarly, \citet{Le-2011-ICA-small} recently showed that adding a regularization
term of the form $\sum_t \sum_j s_3(W_j x^{(t)})$ to a linear auto-encoder with
tied weights, where $s_3$ is a nonlinear convex function, yields an
efficient algorithm for learning \emph{linear ICA}.

\vsD
\subsection{Regularized Auto-Encoders \note{PV, YB}}
\label{sec:ae}
\vsA

Like PCA, auto-encoders were originally seen as a dimensionality
reduction technique and thus used a \emph{bottleneck}, i.e. $d_h < d_x$.
On the other hand, successful uses of sparse coding and RBM approaches tend to favour
\emph{overcomplete} representations,
i.e. $d_h > d_x$. This can allow the auto-encoder
to simply duplicate the input in the features, with perfect
reconstruction without having extracted more meaningful features. 
Recent research has demonstrated very successful
alternative ways, called {\em regulrized auto-encoders},  to ``constrain'' the
representation, even when it is overcomplete.
The effect of a bottleneck or of this regularization
is that the auto-encoder cannot reconstruct well everything,
it is trained to reconstruct well the training examples and generalization
means that reconstruction error is also small on test examples.
An interesting justification~\citep{ranzato-08-small} for the sparsity penalty
(or any penalty that restricts in a soft way the volume of hidden
configurations easily accessible by the learner) is that it acts in spirit
like the partition function of RBMs, by making sure that only few input
configurations can have a low reconstruction error.  

Alternatively, one can view the objective of the regularization applied to an auto-encoder as
making the representation as ``constant'' (insensitive) as possible with respect
to changes in input. This view immediately justifies two variants of regularized
auto-encoders described below: contractive auto-encoders
reduce the number of effective degrees of freedom of the representation
(around each point) by making the encoder contractive, i.e., making the
derivative of the encoder small (thus making the hidden
units saturate), while the denoising auto-encoder makes the whole
mapping ``robust'', i.e., insensitive to small random perturbations,
or contractive, making sure that the
reconstruction cannot stay good when moving in most directions around
a training example.

\vsD
\subsubsection{\bf Sparse Auto-Encoders}
\label{sec:sparse-ae}
\vsA

The earliest uses of single-layer auto-encoders for building deep architectures
by stacking them~\citep{Bengio-nips-2006-small} considered the idea of {\em tying}
the encoder weights and decoder weights to restrict capacity as well as the
idea of introducing a form of {\em sparsity regularization}~\citep{ranzato-07-small}.
Sparsity in the representation 
can be achieved by penalizing the hidden
unit biases (making these additive offset parameters 
more negative)~\citep{ranzato-07-small,HonglakL2008-small,Goodfellow2009-short,Larochelle+Bengio-2008-small}
or by directly penalizing
the output of the hidden unit activations (making them closer to their
saturating value at 0)~\citep{ranzato-08-small,Le-ICML2011-small,Zou-Ng-Yu-NIPSwkshop2011}. 
Penalizing the bias runs the danger that the weights could compensate for the bias,
which could hurt numerical optimization.
When directly penalizing the hidden unit outputs, several variants
can be found in the literature, but a clear comparative analysis
is still lacking. Although the L1
penalty (i.e., simply the sum of output elements $h_j$ in the case of sigmoid non-linearity)
would seem the most natural (because of its use in sparse coding), it
is used in few papers involving sparse auto-encoders. A close cousin of
the L1 penalty is the Student-t penalty ($\log(1+h_j^2)$), originally proposed for
sparse coding~\citep{Olshausen+Field-1996}. Several papers
penalize the {\em average} output $\bar{h}_j$ (e.g. over a minibatch), and instead
of pushing it to 0, encourage it to approach a fixed target, either through
a mean-square error penalty, or maybe more sensibly (because $h_j$ behaves like a probability), 
a Kullback-Liebler divergence  with respect to the binomial distribution with
probability $\rho$: 
$-\rho \log \bar{h}_j -(1-\rho)\log(1-\bar{h}_j)+$constant, e.g., with $\rho=0.05$.


\vsC
\subsubsection{\bf Denoising Auto-Encoders \note{PV}}
\label{sec:dae}
\vsA

\citet{VincentPLarochelleH2008-small,Vincent-JMLR-2010-small} proposed altering the training objective
in Eq.~\ref{eq:reconstruction-error} from mere
reconstruction to that of \emph{denoising} an artificially corrupted input,
i.e. learning to reconstruct the clean input from a corrupted version. 
Learning the identity is no longer enough: the learner must capture the structure
of the input distribution in order to optimally undo the effect of the
corruption process, with the reconstruction essentially being a nearby but higher density
point than the corrupted input. Figure~\ref{fig:dae} illustrates that
the Denoising Auto-Encoder (DAE) is learning a reconstruction function that 
corresponds to a vector field pointing towards high-density regions (the manifold
where examples concentrate).

\begin{figure}[h]
\vsD
\centerline{\includegraphics[width=0.95\linewidth]{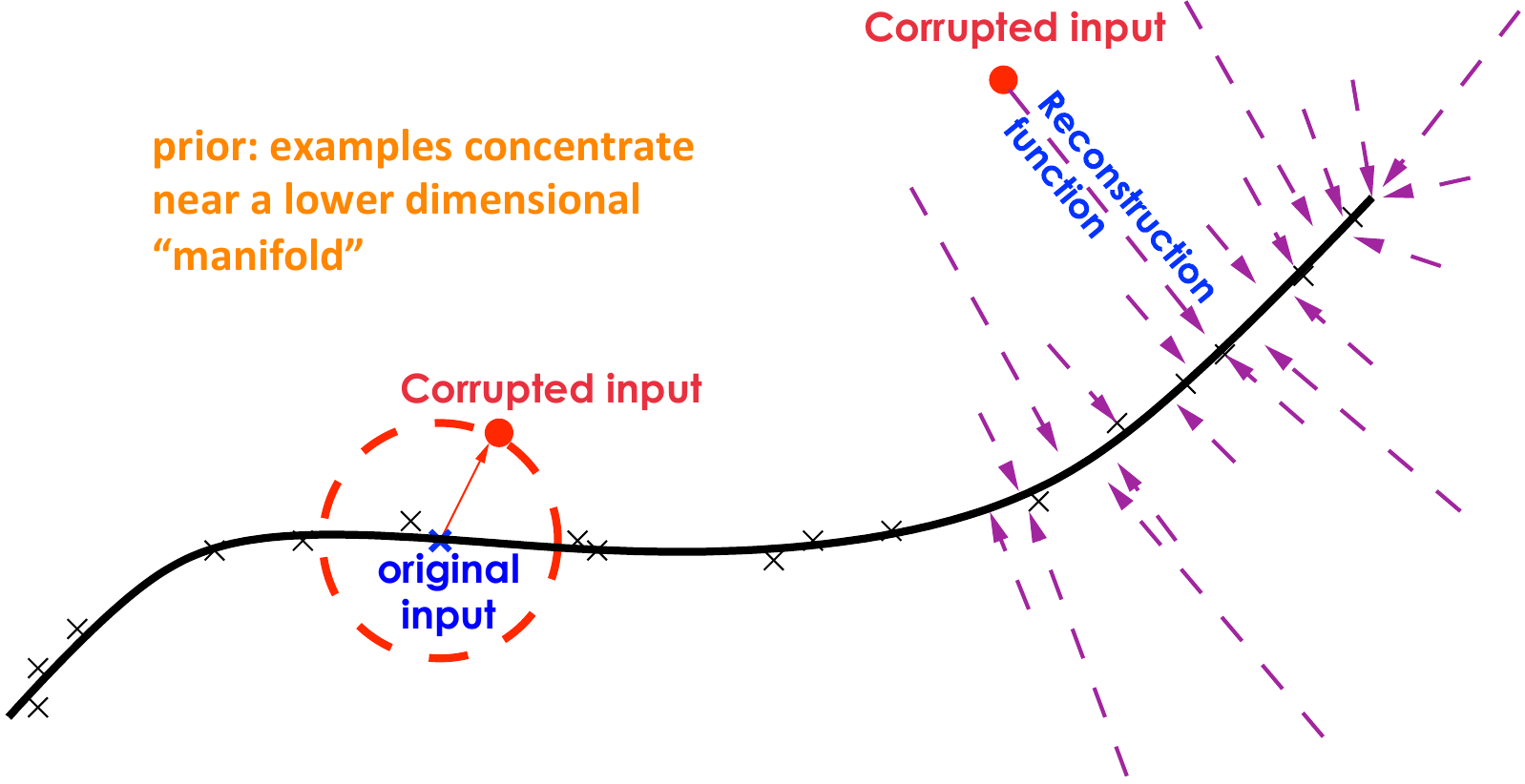}}
\vsD
\caption{\small When data concentrate near a lower-dimensional manifold,
the corruption vector is typically almost orthogonal to the manifold,
and the reconstruction function learns to denoise, map from low-probability
configurations (corrupted inputs) to high-probability ones (original inputs),
creating a vector field aligned with the score (derivative of the
estimated density).}
\label{fig:dae}.
\vsD
\end{figure}

Formally, the objective optimized by a DAE is:

\vsE
\small
\begin{eqnarray}
\mathcal{J}_{\mbox{\tiny DAE}} &=& \sum_t \E{q(\tilde{x}|x^{(t)})}{L(x^{(t)}
  ,g_\theta(f_\theta(\tilde{x})))}
\label{eq:DAE}
\vsG
\end{eqnarray}
\vsD
\normalsize

\noindent where $\E{q(\tilde{x}|x^{(t)})}{\cdot}$ averages over
corrupted examples $\tilde{x}$ drawn from corruption process
$q(\tilde{x}|x^{(t)})$. In practice this is optimized by stochastic
gradient descent, where the stochastic gradient is estimated by drawing one
or a few corrupted versions of $x^{(t)}$ each time $x^{(t)}$ is considered.
Corruptions considered in~\citet{Vincent-JMLR-2010-small} include
additive isotropic Gaussian noise, salt and pepper noise for gray-scale images, and masking noise
(salt or pepper only), e.g., setting some randomly chosen inputs to 0 (independently
per example). Masking noise has been used in most of the simulations.
Qualitatively better features are reported with denoising, resulting
in improved classification, and DAE features performed
similarly or better than  RBM features. \citet{Chen-icml2012}
show that a simpler alternative with a closed form solution can be
obtained when restricting to a \emph{linear}
auto-encoder and have successfully applied it to domain adaptation.

\citet{Vincent-NC-2011-small} relates DAEs
to energy-based probabilistic models:
DAEs basically learn in $r(\tilde{x})-\tilde{x}$
a vector pointing in the direction of the estimated {\em score} 
$\frac{\partial \log p(\tilde{x})}{\partial \tilde{x}}$
(Figure~\ref{fig:dae}).
In the special case of linear reconstruction and squared error,
\citet{Vincent-NC-2011-small} shows that training an affine-sigmoid-affine DAE 
amounts to learning an
energy-based model, whose energy function is very close to that of a
GRBM. Training uses a regularized variant of the \emph{score matching} parameter
estimation
technique~\citep{Hyvarinen-2005-small,HyvarinenA2008,Kingma+LeCun-2010-small} termed
\emph{denoising score matching}~\citep{Vincent-NC-2011-small}. 
\citet{Swersky2010} had shown that training GRBMs with
\emph{score matching} is equivalent to training a regular
auto-encoder with an additional regularization term, while, following
up on the theoretical results in~\citet{Vincent-NC-2011-small}, \citet{Swersky-ICML2011}
showed the practical advantage of denoising  to implement score matching
efficiently. Finally~\citet{Alain+al-arxiv-2012} generalize \citet{Vincent-NC-2011-small}
and prove that DAEs of arbitrary parametrization
with small Gaussian corruption noise are general estimators of the score.

\vsC
\subsubsection{\bf Contractive Auto-Encoders \note{PV}}
\label{sec:CAE}
\vsA

Contractive Auto-Encoders (CAE), proposed by~\citet{Rifai+al-2011-small}, follow up on
Denoising Auto-Encoders (DAE) and share a similar motivation of learning robust
representations. CAEs achieve this by adding an analytic \emph{contractive penalty}
to Eq.~\ref{eq:reconstruction-error}: the Frobenius norm of the
encoder's Jacobian, and results in penalizing the \emph{sensitivity} of learned
features to infinitesimal  input variations.
Let $J(x) = \frac{\partial f_\theta}{\partial x}(x)$ the Jacobian matrix of the
encoder at $x$. The CAE's training objective is 

\vsE
\small
\begin{eqnarray}
\mathcal{J}_{\mbox{\tiny CAE}} &=& \sum_t L(x^{(t)}
  ,g_\theta(f_\theta(x^{(t)}))) + \lambda \left\| J(x^{(t)}) \right\|^2_F
\label{eq:CAE}
\vsE
\end{eqnarray}
\vsD
\normalsize

\noindent where $\lambda$ is a hyper-parameter controlling the strength of the regularization.
For an affine sigmoid encoder, the contractive penalty term is easy to compute:

\vsF
\addtolength{\jot}{0.5mm}
\small
\begin{eqnarray}
J_j(x) &=& f_\theta(x)_j (1-f_\theta(x)_j) W_j \nonumber\\
\left\| J(x) \right\|^2_F 
  &=& \sum_j (f_\theta(x)_j (1-f_\theta(x)_j))^2 \|W_j\|^2
\vsD
\label{eq:Jacobian}
\end{eqnarray}
\vsC
\normalsize
\addtolength{\jot}{-0.5mm}

\noindent There are at least three notable differences with DAEs, which may be partly
responsible for the better performance that CAE features seem to empirically demonstrate: (a) the sensitivity
of the \emph{features} is penalized\footnote{i.e., the robustness of the
representation is encouraged.} rather than that of the 
\emph{reconstruction}; (b) the penalty is analytic rather than stochastic: an efficiently computable
expression replaces what might otherwise require $d_x$ corrupted samples to
size up (i.e. the sensitivity in $d_x$ directions); (c) a hyper-parameter
$\lambda$ allows
a fine control of the trade-off between reconstruction and robustness (while the two are mingled in a DAE).
Note however that there is a tight connection between the DAE and the CAE: as shown
in~\citep{Alain+al-arxiv-2012} a
DAE with small corruption noise
can be seen (through a Taylor expansion)
as a type of contractive auto-encoder where the contractive penalty
is on the whole reconstruction function rather than just on the encoder\footnote{but
note that in the CAE, the decoder weights are tied to the encoder weights, to avoid
degenerate solutions, and this should also make the decoder contractive.}.

A potential disadvantage of the CAE's analytic penalty is that it amounts to
only encouraging robustness to \emph{infinitesimal} input variations.  
This is remedied in \citet{Salah+al-2011-small} with
the CAE+H, that penalizes all higher order derivatives, in an efficient stochastic manner,
by adding a term that encourages $J(x)$ and $J(x+\epsilon)$ to be close:

\vsF
\small
\begin{eqnarray}
\mathcal{J}_{\mbox{\tiny CAE+H}} &=& \sum_t L(x^{(t)}, g_\theta(x^{(t)})) 
                  + \lambda \left\| J(x^{(t)}) \right\|^2_F \nonumber \\
              & & + \gamma \E{\epsilon}{\left\| J(x) -
                J(x+\epsilon) \right\|^2_F}
\label{eq:CAE+H}
\vsD
\end{eqnarray}
\normalsize
where $\epsilon \sim \mathcal{N}(0, \sigma^2 I)$, and $\gamma$ is the
associated regularization strength hyper-parameter. 

The DAE and CAE have been successfully used to win the final phase of the Unsupervised and Transfer
Learning Challenge~\citep{UTLC+LISA-2011-small}.
The representation learned by the CAE tends to be {\em saturated
rather than sparse}, i.e., most hidden units are near the extremes
of their range (e.g. 0 or 1), and their derivative $\frac{\partial h_i(x)}{\partial x}$
is near 0. The non-saturated units are few and sensitive to the inputs, with their
associated filters (weight vectors) together forming a basis explaining the local changes around $x$,
as discussed in Section~\ref{sec:manifold-coding}. Another way to get
saturated (nearly binary) units is
{\em semantic hashing}~\citep{Salakhutdinov+Hinton2007-small}.

\vsD
\subsubsection{\bf Predictive Sparse Decomposition \note{YB}}
\label{sec:PSD}
\vsA

Sparse coding~\citep{Olshausen+Field-1996} may be viewed as a kind of
auto-encoder that uses a linear decoder with a squared reconstruction
error, but whose non-parametric \emph{encoder} $f_\theta$ performs the comparatively non-trivial and relatively costly
iterative minimization of Eq.~\ref{eq:sparse-coding-map}.
A practically successful variant of sparse coding and auto-encoders, named {\em Predictive Sparse Decomposition} 
or PSD~\citep{koray-psd-08-small} replaces that costly and highly non-linear encoding step by a fast non-iterative approximation
during recognition (computing the learned features).
PSD has been applied to object recognition in images and 
video~\citep{Koray-08-small,koray-nips-10-small,Jarrett-ICCV2009-small},
but also to audio~\citep{henaff-ismir-11-small}, mostly within 
the framework of multi-stage convolutional deep architectures
(Section~\ref{sec:convol}). 
The main idea can be summarized
by the following equation for the training criterion, which is simultaneously
optimized with respect to hidden codes (representation) $h^{(t)}$ and with
respect to parameters $(W,\alpha)$:

\vsF
\small
\begin{equation}
  \mathcal{J}_{\mbox{\tiny PSD}} = \sum_t \lambda \|h^{(t)}\|_1 + \|x^{(t)} - W h^{(t)}\|^2_2 + \|h^{(t)} - f_\alpha(x^{(t)})\|^2_2
\label{eq:PSD}
\vsD
\end{equation}
\normalsize
where $x^{(t)}$ is the input vector for example $t$, $h^{(t)}$ is the
optimized hidden code for that example, and $f_\alpha(\cdot)$ is
the encoding function, the simplest variant being

\vsD
\small
\begin{equation}
  f_\alpha(x^{(t)}) = \tanh(b+W^T x^{(t)})
\label{eq:PSDenc}
\vsC
\end{equation}
\vsE
\normalsize

\noindent where encoding weights are the transpose of decoding
weights. Many variants have been proposed, including
the use of a shrinkage operation instead of the 
hyperbolic tangent~\citep{koray-nips-10-small}.
Note how the L1 penalty on $h$ tends to make them sparse, and
how this is the same criterion as sparse coding with
dictionary learning (Eq.~\ref{eq:sparse-coding-cost}) except for the additional constraint that
one should be able to approximate the sparse codes $h$ with
a parametrized encoder $f_\alpha(x)$.
One can thus view PSD as an approximation to sparse coding,
where we obtain a fast approximate encoder.
Once PSD is trained, object
representations $f_\alpha(x)$ are used to feed a classifier. They are computed
quickly and can be further fine-tuned:
the encoder can be viewed as one stage or one layer of
a trainable multi-stage system such as a feedforward neural network.

PSD can also be seen as a kind of auto-encoder 
where the codes $h$ are given some freedom that can help
to further improve reconstruction. One can also view the encoding
penalty added on top of sparse coding as a kind of regularizer that
forces the sparse codes to be nearly computable by a smooth and
efficient encoder. This is in contrast with the codes obtained
by complete optimization of the sparse coding criterion, which 
are highly non-smooth or even non-differentiable, a problem
that motivated
other approaches to smooth the inferred codes of sparse coding~\citep{Bradley+Bagnell-2009-small},
so a sparse coding stage could be jointly optimized along with
following stages of a deep architecture.



\vsE
\section{Representation Learning as Manifold Learning \note{PV}}
\label{sec:manifold}
\vsA

Another important perspective on representation learning is based on the geometric notion
of manifold. Its premise is the 
\emph{manifold
  hypothesis}, according to which 
real-world data presented in high dimensional spaces are expected to
concentrate in the vicinity of a manifold $\mathcal{M}$ of much lower
dimensionality $d_\mathcal{M}$, embedded in high dimensional input
space $\R^{d_x}$. This prior seems particularly well suited 
for AI tasks such as those involving images, sounds or text, for
which most uniformly sampled input configurations are unlike natural
stimuli.
As soon as there is a notion of ``representation'' then
one can think of a manifold by considering the \emph{variations} in input space,
which are captured by or reflected (by corresponding changes)
in the learned representation. To first approximation,
some directions are well preserved (the tangent directions of the
manifold) while others aren't (directions orthogonal to the manifolds).
With this perspective, the primary unsupervised learning task is then seen as modeling the
structure of the data-supporting manifold\footnote{Actually,
  data points need not strictly lie
  on the ``manifold'', but the probability density is expected to fall off sharply as one 
  moves away from it, and it may actually be constituted of several
  possibly disconnected manifolds with different intrinsic
  dimensionality.}.
The associated \emph{representation} being learned can be associated with
an intrinsic coordinate system on the embedded manifold.
The archetypal manifold modeling algorithm is, not surprisingly, 
also the archetypal low dimensional representation learning algorithm:
Principal Component Analysis, which models a
\emph{linear} manifold. It was initially devised with the objective of 
finding the closest linear manifold to a cloud of data points. The principal components, i.e. the
\emph{representation} $f_\theta(x)$ that PCA yields for an input point $x$,
uniquely locates its projection on that manifold: it corresponds to
intrinsic coordinates on the manifold. 
Data manifold for complex real world domains are however expected to be
\emph{strongly non-linear}. Their modeling is sometimes approached as patchworks of locally linear
tangent spaces~\citep{Vincent-Bengio-2003-short,Brand2003-small}.
The large majority of algorithms built on this geometric perspective
adopt a non-parametric approach, based on a training set nearest neighbor
graph~\citep{Scholkopf98,
  Roweis2000-lle-small,Tenenbaum2000-isomap,Brand2003-small,Belkin+Niyogi-2003,Donoho+Carrie-03,Weinberger04a-small,SNE-nips15-small,VanDerMaaten08-small}. In
these non-parametric approaches, each high-dimensional training
point has its own set of free low-dimensional
\emph{embedding} coordinates, which are optimized so that certain properties of the neighborhood
graph computed in original high dimensional input space are best preserved. 
These methods however do not directly learn a parametrized feature extraction
function $f_\theta(x)$ applicable to new test points\footnote{For several of
  these techniques, representations for new points can be computed using the Nystr\"om
approximation as has been proposed as an extension in~\citep{Bengio-nips2003-small}, but 
this remains cumbersome and computationally expensive.}, which
seriously limits their use as feature extractors, except in a transductive setting. 
Comparatively few non-linear manifold learning methods have been proposed, that learn a
\emph{parametric map} that can directly compute a representation for new points; we will focus on these.

\vsE
\subsection{Learning a parametric mapping based on a neighborhood graph}
\label{sec:neighbors}
\vsB

Some of the above non-parametric manifold learning algorithms
can be modified to learn a \emph{parametric
  mapping} $f_\theta$, i.e., applicable to new points:
instead of having \emph{free} low-dimensional
embedding coordinate ``parameters'' for each training point, these
coordinates are obtained through an explicitly parametrized function,
as with the parametric variant~\citep{VanDerMaaten09-small} 
of t-SNE~\citep{VanDerMaaten08-small}.

Instead, Semi-Supervised Embedding~\citep{WestonJ2008-small}
learns a direct encoding while taking
into account the manifold hypothesis through a neighborhood graph.
A parametrized neural network architecture simultaneously learns a manifold
embedding and a classifier. The training criterion encourages 
training set neigbhors to have similar representations.

The reduced and tightly controlled number of free parameters in such parametric
methods, compared to their pure non-parametric counterparts, forces 
models to generalize the manifold shape
non-locally~\citep{Bengio-Larochelle-NLMP-NIPS-2006-short}, which
can translate into better features and final
performance~\citep{VanDerMaaten08-small}.
However, basing the modeling of manifolds on training set neighborhood
relationships might be risky statistically in high dimensional spaces
(sparsely populated due to the curse of dimensionality) as e.g. most
Euclidean nearest neighbors risk having too little in common semantically. 
The nearest neighbor graph is simply not enough densely populated to map
out satisfyingly the wrinkles of the target manifold~\citep{Bengio+Monperrus-2005-short,Bengio-Larochelle-NLMP-NIPS-2006-short,Bengio+Lecun-chapter2007}. 
It
can also become problematic computationally to consider all
pairs of data points\footnote{Even if pairs are picked stochastically, many
  must be considered before obtaining one that weighs significantly on the
  optimization objective.}, which scales quadratically with training set
size.

\vsD
\subsection{Learning to represent non-linear manifolds}
\label{sec:manifold-coding}
\vsA

Can we learn a manifold without requiring nearest neighbor searches? Yes, for
example, with regularized auto-encoders or PCA. In PCA, 
the sensitivity of the extracted components (the code) to input changes 
is the same regardless of position $x$. The tangent space is the same
everywhere along the linear manifold. By contrast, for a {\em non-linear
manifold}, the tangent of the manifold changes as we
move on the manifold, as illustrated in Figure~\ref{fig:manifold-sampling}. 
In non-linear representation-learning algorithms it is convenient
to think about the {\em local variations} in the representation as
the input $x$ is varied {\em on the manifold}, i.e., as we move among
high-probability configurations.
As we discuss below, the first derivative of 
the encoder therefore specifies the shape
of the manifold (its tangent plane) around an example $x$ lying on it.
If the density was really concentrated on the manifold, 
and the encoder had captured that, we would find the encoder derivatives
to be non-zero only in the directions spanned by the tangent plane.
 
Let us consider sparse coding in this light: parameter matrix $W$ may be
interpreted as a dictionary of input directions from which a \emph{different subset} will be
picked to model the local tangent space at an $x$ on the manifold. That subset corresponds
to the active, i.e. non-zero, features for input $x$. Non-zero component $h_i$
will be sensitive to small changes of the input in the direction of the 
associated weight vector $W_{:,i}$, whereas inactive features are more likely to be stuck
at 0 until a significant displacement has taken place in input space.

The \emph{Local Coordinate Coding} (LCC) algorithm~\citep{Yu+al-NIPS09-small} is
very similar to sparse coding, but is explicitly derived from a manifold
perspective. Using the same notation as that of sparse coding in
Eq.~\ref{eq:sparse-coding-map}, LCC
replaces regularization term $\|h^{(t)}\|_1 = \sum_j |h^{(t)}_j|$ yielding objective

\small
\vsF
\begin{equation}
  \mathcal{J}_{\mbox{\tiny LCC}} = \sum_t \left(\|x^{(t)} - W h^{(t)}\|^2_2 + \lambda \sum_j |h^{(t)}_j| \| W_{:,j} -
    x^{(t)} \|^{1+p} \right)
\label{eq:sparse-coding}
\vsD
\end{equation}
\normalsize
\vsC

\noindent This is identical to sparse coding when $p=-1$, but with larger $p$ it 
encourages the active \emph{anchor points} for $x^{(t)}$ (i.e. the codebook vectors $W_{:,j}$ 
with non-negligible $|h^{(t)}_j|$ that are combined to reconstruct $x^{(t)}$) to be not too far from $x^{(t)}$, hence the
\emph{local} aspect of the algorithm. An important theoretical contribution
of~\citet{Yu+al-NIPS09-small} is to show that
that any Lipschitz-smooth function $\phi: \mathcal{M} \rightarrow \R$
defined on a smooth
nonlinear manifold $\mathcal{M}$ embedded in $\R^{d_x}$
can be well approximated by a globally \emph{linear function} with
respect to the resulting coding scheme (i.e. linear in $h$), where the accuracy of the
approximation and required number $d_h$ of anchor points depend on $d_\mathcal{M}$ rather than $d_x$.
This result has been further extended with
the use of local tangent directions~\citep{yu2010tangents-small}, 
as well as to multiple layers~\citep{Lin+Tong+Yu-DeepCodingNetwork-2010-small}.


Let us now consider the efficient non-iterative ``feed-forward'' encoders
$f_\theta$, used by PSD and the auto-encoders reviewed in
Section~\ref{sec:ae}, that are in
the form of Eq. \ref{eq:AE} or \ref{eq:PSDenc}.
The computed representation for $x$ will be only significantly sensitive to
input space directions associated with non-saturated hidden
units (see e.g. Eq.~\ref{eq:Jacobian} for the Jacobian of a sigmoid
layer). These directions to which the representation is
significantly sensitive, like in the case of PCA or sparse coding, may be viewed as
spanning the tangent space of the manifold at training point $x$.

\begin{figure}[h]
\vsD
\centerline{\includegraphics[width=0.69\linewidth]{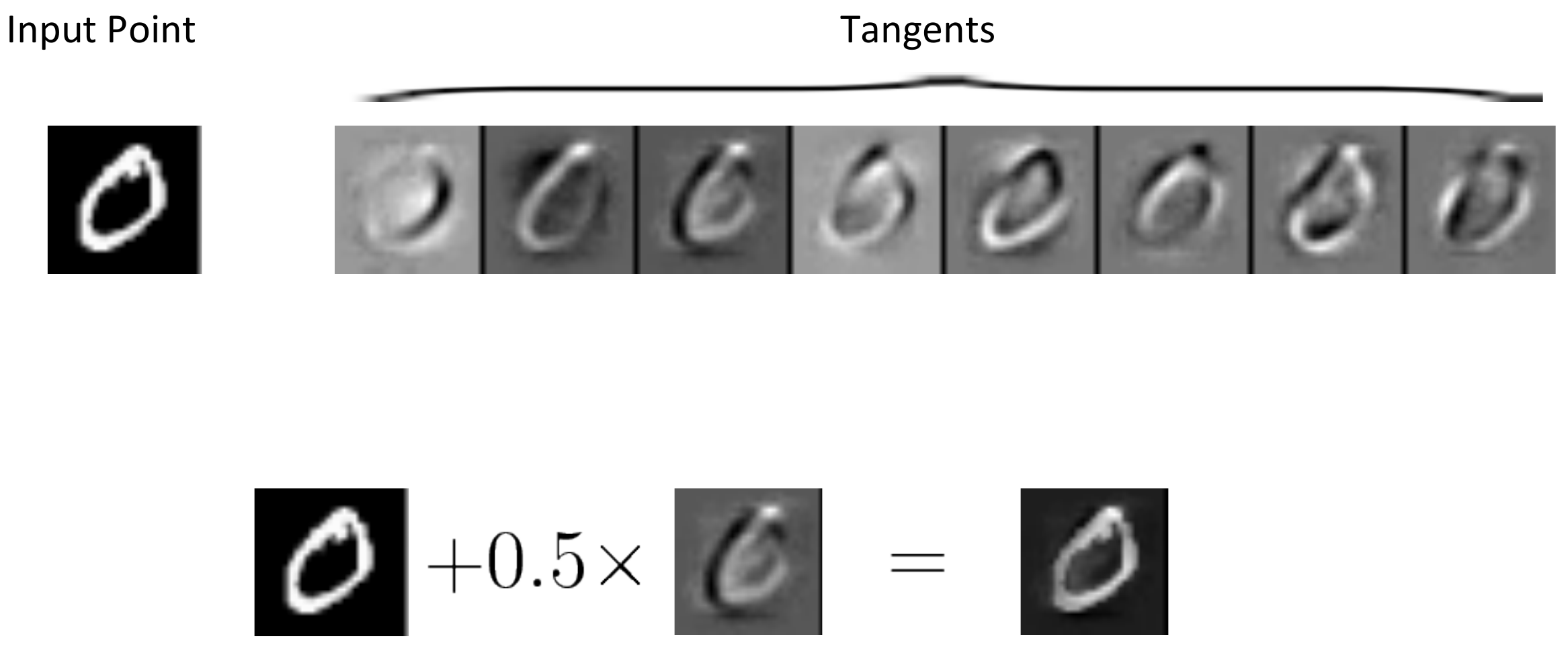}}
\vsD
\caption{\small The tangent vectors to the high-density manifold
as estimated by a Contractive Auto-Encoder~\citep{Rifai+al-2011-small}.
The original input is shown on the top left. Each tangent vector (images
on right side of first row)
corresponds to a plausible additive deformation of the original input,
as illustrated on the second row, where a bit of the 3rd singular
vector is added to the original, to form a translated and deformed image. Unlike
in PCA, the tangent vectors are different for different inputs, because
the estimated manifold is highly non-linear.
} \label{fig:tangent-vectors}.
\vsF
\end{figure}

\citet{Rifai+al-2011-small} empirically analyze in this light the singular value
spectrum of the Jacobian (derivatives of representation vector with respect
to input vector) 
of a trained CAE. Here the SVD provides an ordered
orthonormal basis of most sensitive directions. The spectrum is sharply decreasing,
indicating a relatively small number of significantly sensitive
directions. This is taken as empirical evidence that the CAE indeed modeled
the tangent space of a low-dimensional manifold. The leading singular
vectors form a basis for the tangent plane of the estimated manifold,
as illustrated in Figure~\ref{fig:tangent-vectors}.
The CAE criterion is
believed to achieve this thanks to its two opposing terms: the isotropic
contractive penalty, that encourages the representation to be equally
insensitive to changes in any input directions, and the reconstruction
term, that pushes different training points (in particular neighbors) to
have a different representation (so they may be reconstructed accurately),
thus counteracting the isotropic contractive pressure only in directions
tangent to the manifold.

Analyzing learned representations through the lens of the
spectrum of the Jacobian and relating it to the notion of tangent space of a manifold
is feasible, whenever the mapping is differentiable, and regardless of how it was
learned, whether as direct encoding (as in auto-encoder variants), or
derived from latent variable inference (as in sparse coding or RBMs). 
Exact low dimensional manifold models (like PCA) would yield non-zero
singular values associated to directions along the manifold, and exact zeros
for directions orthogonal to the manifold. But in smooth
models like the CAE or the RBM we will instead have
large versus relatively small singular values (as opposed to non-zero versus
exactly zero).

\vsD
\subsection{Leveraging the modeled tangent spaces}
\label{sec:leveraging-manifold}
\vsA

The local tangent space, at a point along the manifold, can be thought of
capturing \emph{locally} valid transformations that were prominent in the
training data.
For example \citet{Dauphin-et-al-NIPS2011-small} examine the
tangent directions extracted with an SVD of the Jacobian of 
CAEs trained on digits, images, or text-document data: they appear to correspond to
small translations or rotations for images or digits, and to
substitutions of words within a same theme for documents.
Such very local transformations along a data manifold are not expected to change class
identity. To build their Manifold Tangent Classifier (MTC), \citet{Dauphin-et-al-NIPS2011-small} then apply techniques such as
\emph{tangent distance}~\citep{Simard93-small} and \emph{tangent
  propagation}~\citep{Simard92-short}, that were initially developed to build
classifiers that are insensitive to input deformations provided as
prior domain knowledge. Now these techniques are applied
using the local leading tangent directions extracted
by a CAE, i.e. not using \emph{any} prior domain knowledge
(except the broad prior about the existence of a manifold).
This approach set a new record for MNIST digit
classification among prior-knowledge free approaches\footnote{It yielded 0.81\% error
  rate using the full MNIST training set, with no prior deformations, and no
convolution.}.

\vsC
\section{Connections between Probabilistic and Direct Encoding models \note{YB}}
\label{sec:connections}
\vsA

The standard likelihood framework for probabilistic models decomposes
the training criterion for models with parameters $\theta$
in two parts: the log-likelihood $\log P(x|\theta)$ 
(or $\log P(x|h,\theta)$ with latent variables $h$), and the prior
$\log P(\theta)$ (or $\log P(h|\theta) + \log P(\theta)$ with latent
variables).

\vsD
\subsection{PSD: a probabilistic interpretation}
\vsA

In the case of the PSD algorithm, a connection can be
made between the above standard probabilistic view and the direct encoding computation
graph. The probabilistic model of PSD is the same directed generative
model $P(x|h)$ of sparse coding (Section~\ref{sec:sparse-coding}), which
only accounts for the decoder. The encoder is viewed as an approximate
inference mechanism to guess $P(h|x)$ and initialize a MAP iterative
inference (where the sparse prior $P(h)$ is taken into account). However,
in PSD, the encoder is {\em trained jointly with the decoder},
rather than simply taking the end result of iterative inference as a target
to approximate. An interesting view\footnote{suggested by Ian Goodfellow,
  personal communication} to reconcile these facts is that the encoder is a
{\em parametric approximation for the MAP solution of a variational lower bound
on the joint log-likelihood}. When MAP learning is viewed as a special case of
variational learning (where the approximation of the joint log-likelihood
is with a dirac distribution located at the MAP solution), the variational
recipe tells us to simultaneously improve the likelihood (reduce reconstruction
error) and improve the variational approximation (reduce the discrepancy
between the encoder output and the latent variable value).
Hence PSD sits at the intersection of probabilistic models (with latent
variables) and direct encoding methods (which directly parametrize
the mapping from input to representation). RBMs also sit at the intersection
because their particular parametrization includes an explicit
mapping from input to representation, thanks to the restricted connectivity
between hidden units. However, this nice property does not extend to their
natural deep generalizations, i.e., 
Deep Boltzmann Machines, discussed in Section~\ref{sec:DBM}.

\vsD
\subsection{Regularized Auto-Encoders Capture Local Structure of the Density}
\vsA

Can we also say something about the probabilistic interpretation
of regularized auto-encoders?
Their training criterion does not fit the standard likelihood framework
because this would involve a {\em data-dependent} ``prior''.
An interesting hypothesis emerges to answer that question, out of recent
theoretical
results~\citep{Vincent-NC-2011-small,Alain+al-arxiv-2012}: the
training criterion of regularized auto-encoders, instead of being a form of maximum likelihood,
corresponds to a different inductive principle, such as {\em score
  matching}. The score matching connection is discussed in
Section~\ref{sec:dae} and has been shown for a particular parametrization
of DAE and equivalent Gaussian
RBM~\citep{Vincent-NC-2011-small}. The work
in~\cite{Alain+al-arxiv-2012} generalizes this idea to a broader
class of parametrizations (arbitrary encoders and decoders), and
shows that by regularizing the auto-encoder so that it be contractive,
one obtains that the reconstruction function
and its derivative {\em estimate first and second derivatives} of the underlying
data-generative density. This view can be exploited
to successfully {\em sample} from auto-encoders, as shown
in~\cite{Rifai-icml2012-small,Bengio-arxiv-moments-2012}. The proposed
sampling algorithms are MCMCs similar to Langevin MCMC, using
not just the estimated first derivative of the density but also
the estimated manifold tangents so as to stay close to manifolds
of high density.

\begin{figure}[h]
\vsD
\centerline{\includegraphics[width=0.85\linewidth]{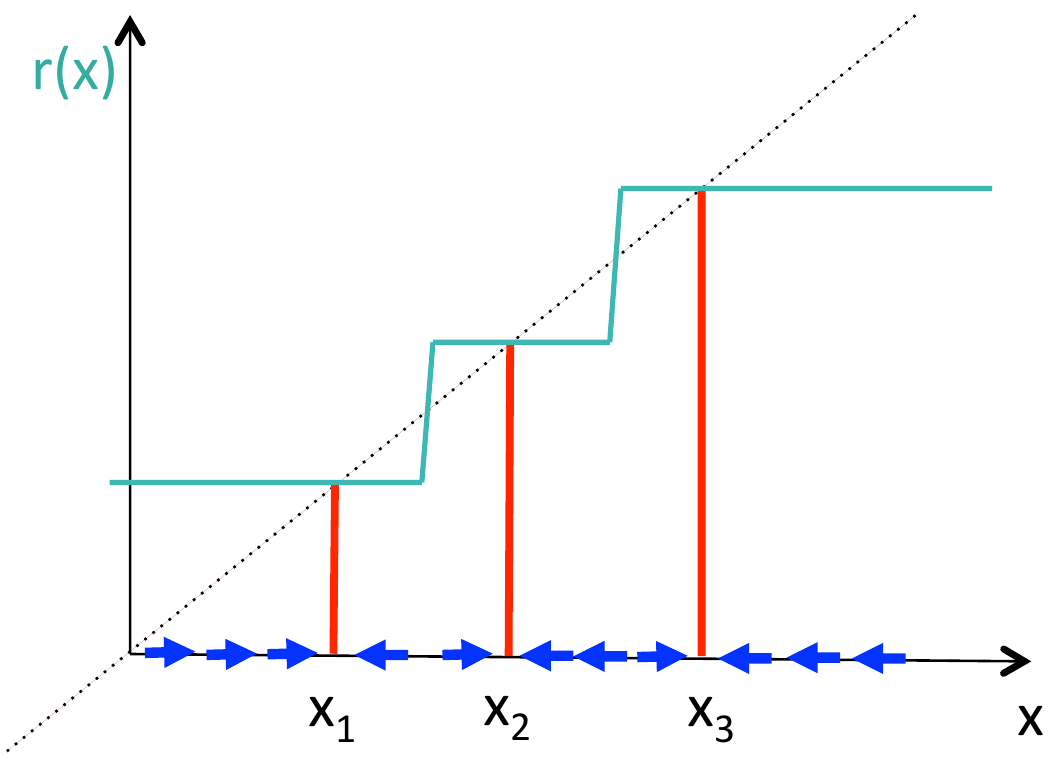}}
\vsD
\caption{\small Reconstruction function $r(x)$ (green)
learned by a high-capacity autoencoder
on 1-dimensional input, minimizing
reconstruction error {\em at training examples} $x^{(t)}$
($r(x^{(t)})$ in red) while
trying to be as constant as possible otherwise. The dotted line is
the identity reconstruction (which might be obtained without the regularizer).
The blue arrows shows the vector field of $r(x)-x$ pointing towards
high density peaks estimated by the model, and estimating
the score (log-density derivative).
} \label{fig:1D-autoencoder}.
\vsD
\end{figure}

This interpretation connects well with the geometric perspective introduced
in Section~\ref{sec:manifold}. The regularization effects (e.g., due to a
sparsity regularizer, a contractive regularizer, or the denoising
criterion) asks the learned representation to be as insensitive as possible
to the input, while minimizing reconstruction error on the training
examples forces the representation to contain just enough information to
distinguish them. The solution is that variations along the high-density
manifolds are preserved while other variations are compressed:
the reconstruction function should be as constant as possible while
reproducing training examples, i.e., points near a training example
should be mapped to that training example
(Figure~\ref{fig:1D-autoencoder}). 
The reconstruction function should map an input towards the nearest point
manifold, i.e., the difference between reconstruction and input is a vector
aligned with the estimated score (the derivative of the log-density with
respect to the input). The score can be zero on the manifold (where reconstruction
error is also zero), at local maxima of the log-density, but it can also be zero at 
local minima. It means that we cannot equate low reconstruction error with high
estimated probability. The second derivatives of the log-density
corresponds to the first derivatives of the reconstruction function, and on the
manifold (where the first derivative is 0), they indicate the tangent directions of
the manifold (where the first derivative remains near 0).

\begin{figure}[h]
\vsD
\hspace*{2mm}\centerline{\includegraphics[width=1.1\linewidth]{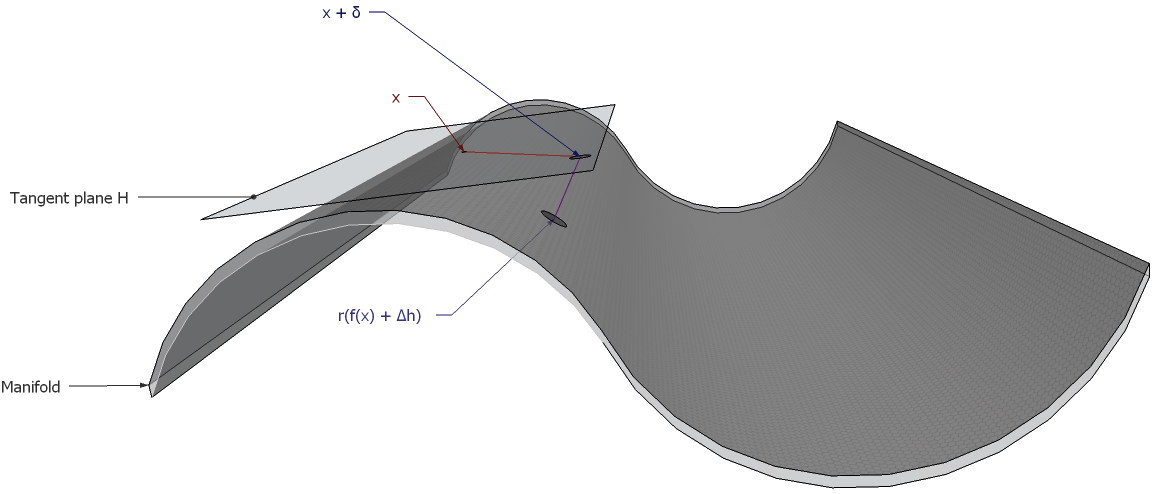}}
\vsC
\caption{\small Sampling from regularized
auto-encoders~\citep{Rifai-icml2012-small,Bengio-arxiv-moments-2012}:
each MCMC step adds to current state $x$ the noise $\delta$ mostly in the directions
of the estimated manifold tangent plane $H$ and projects back towards
the manifold (high-density regions) by performing a reconstruction 
step.
} \label{fig:manifold-sampling}.
\vsD
\end{figure}

As illustrated in Figure~\ref{fig:manifold-sampling},
the basic idea of the auto-encoder sampling algorithms
in~\cite{Rifai-icml2012-small,Bengio-arxiv-moments-2012} is to make
MCMC moves where one (a) moves toward the manifold by following the
density gradient (i.e., applying a reconstruction) and (b) adds noise
in the directions of the leading singular vectors of the reconstruction
(or encoder) Jacobian, corresponding to those associated with smallest
second derivative of the log-density.

\vsD
\subsection{Learning Approximate Inference}
\vsA

Let us now consider from closer how a representation is computed in
probabilistic models with latent variables, when iterative inference is
required. There is a computation graph (possibly with random number
generation in some of the nodes, in the case of MCMC) that maps inputs to
representation, and in the case of deterministic inference (e.g., MAP
inference or variational inference), that function could be optimized
directly. This is a way to generalize PSD that has been explored in recent
work on probabilistic models at the intersection of inference and
learning~\citep{Bradley+Bagnell-2009-small,Gregor+LeCun-ICML2010-small,Grubb+Bagnell-2010-small,Salakhutdinov+Larochelle-2010-small,Stoyanov2011-small,eisner-2012-icmlw}, where a central
idea is that instead of using a {\em generic} inference mechanism,
one can use one that is {\em learned} and is more efficient, taking advantage
of the specifics of the type of data on which it is applied.

\vsD
\subsection{Sampling Challenges}
\label{sec:sampling-challenge}
\vsA

A troubling challenge with many probabilistic models with latent variables
like most Boltzmann machine variants is that good MCMC sampling is required
as part of the learning procedure, but that {\em sampling becomes extremely
inefficient} (or unreliable) as training progresses because the modes of the learned distribution
become sharper, making {\em mixing between modes very slow}. Whereas initially
during training a learner assigns mass almost uniformly, as training progresses,
its entropy decreases, approaching the entropy of the target distribution as
more examples and more computation are provided. According to 
our Manifold and Natural Clustering priors of Section~\ref{sec:priors},
the target distribution has sharp modes (manifolds) separated by extremely
low density areas. Mixing then becomes more difficult because MCMC methods,
by their very nature, tend to make small steps to nearby high-probability
configurations. This is illustrated in Figure~\ref{fig:mixing-issue}.

\begin{figure}[h]
\vsD
\centerline{\includegraphics[width=0.7\linewidth]{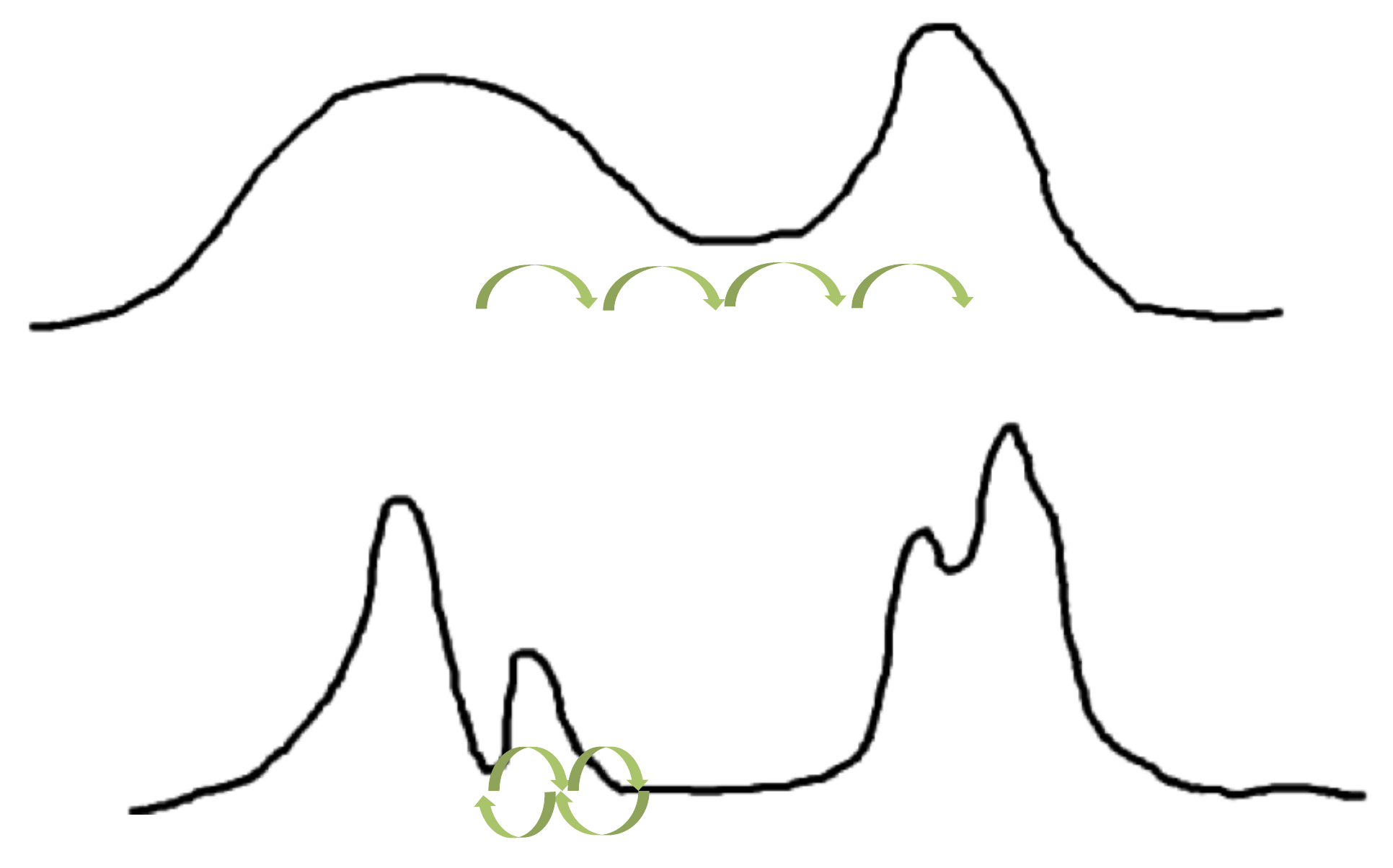}}
\vsD
\caption{\small Top: early during training, MCMC mixes easily between
modes because the estimated
distribution has high entropy and puts enough mass everywhere for small-steps
movements (MCMC) to go from mode to mode. Bottom: later on, training relying
on good mixing can stall because estimated modes are separated by wide
low-density deserts.} \label{fig:mixing-issue}.
\vsD
\end{figure}

\citet{Bengio-et-al-ICML2013} suggest that deep representations could
help mixing between such well separated modes, based on both theoretical
arguments and on empirical evidence. The idea is that if higher-level
representations disentangle better the underlying abstract factors, then
small steps in this abstract space (e.g., swapping from one category to
another) can easily be done by MCMC. The high-level representations can
then be mapped back to the input space in order to obtain input-level
samples, as in the Deep Belief Networks (DBN) sampling
algorithm~\citep{Hinton06}.  This has been demonstrated both with DBNs and
with the newly proposed algorithm for sampling from contracting and
denoising
auto-encoders~\citep{Rifai-icml2012-small,Bengio-arxiv-moments-2012}. This
observation alone does not suffice to solve the problem of training a DBN
or a DBM, but it may provide a crucial ingredient, and it makes it possible
to consider successfully sampling from deep models trained by procedures
that do not require an MCMC, like the stacked regularized auto-encoders
used in~\citet{Rifai-icml2012-small}.

\vsD
\subsection{Evaluating and Monitoring Performance}
\vsA

It is always possible to evaluate a feature learning algorithm in terms of
its usefulness with respect to a particular task (e.g. object
classification), with a predictor that is fed or initialized with the
learned features. In practice, we do this by saving the features learned
(e.g. at regular intervals during training, to perform early stopping)
and training a cheap classifier on top (such as a linear classifier).
However, training the final classifier can be a substantial computational overhead
(e.g., supervised fine-tuning a deep neural network takes usually more training
iterations than the feature learning itself), so we may want to avoid having
to train a classifier for every training iteration of the unsupervised
learner and every hyper-parameter setting.
More importantly this may give an incomplete evaluation of the features (what would happen for
other tasks?).  All these issues motivate the use of methods to monitor and evaluate
purely unsupervised performance.  This is rather easy with all the
auto-encoder variants (with some caution outlined below) and rather
difficult with the undirected graphical models such as the RBM and
Boltzmann machines.

For auto-encoder and sparse coding variants, test set reconstruction error
can readily be computed, but by itself may be misleading because {\em
  larger capacity} (e.g., more features, more training time) tends to
systematically lead to lower reconstruction error, even on the test
set. Hence it cannot be used reliably for selecting most hyper-parameters.
On the other hand, {\em denoising} reconstruction error is clearly immune
to this problem, so that solves the problem for DAEs.
Based on the connection between DAEs and CAEs uncovered in
~\citet{Bengio-arxiv-moments-2012,Alain+al-arxiv-2012}, this immunity
can be extended to DAEs, but not to the hyper-parameter controlling
the amount of noise or of contraction.


For RBMs and some (not too deep) Boltzmann machines, one option is the use
of Annealed Importance Sampling~\citep{MurraySal09-small} in order to estimate
the partition function (and thus the test log-likelihood). Note that this
estimator can have high variance and
that it becomes less reliable (variance becomes too large)
as the model becomes more interesting, with
larger weights, more non-linearity, sharper modes and a sharper probability
density function (see our previous discussion in Section~\ref{sec:sampling-challenge}).  
Another interesting and recently proposed option for
RBMs is to {\em track} the partition function during training~\citep{Desjardins+al-NIPS2011},
which could be useful for early stopping and reducing the cost of ordinary AIS.
For toy RBMs (e.g., 25 hidden units or less, or 25
inputs or less), the exact log-likelihood can also be computed
analytically, and this can be a good way to debug and verify some
properties of interest. 

\vsD
\section{Global Training of Deep Models}
\label{sec:global}
\vsA

One of the most interesting challenges raised by deep architectures is: {\em how should
we jointly train all the levels}? In the previous section
and in Section~\ref{sec:stacking} we have only discussed how
single-layer models could be combined to form a deep model.
Here we consider joint training of all the levels and the difficulties
that may arise.

\vsD
\subsection{The Challenge of Training Deep Architectures \note{YB}}
\vsA

Higher-level abstraction means more non-linearity. It means that two nearby
input configurations may be interpreted very differently
because a few surface details change the underlying semantics, whereas most
other changes in the surface details would not change the underlying semantics. The
representations associated with input manifolds may be complex because
the mapping from input to representation may have to unfold and distort input manifolds that generally
have complicated shapes into spaces where distributions are much simpler,
where relations between factors are simpler, maybe even linear or involving
many (conditional) independencies. Our expectation is that modeling the
joint distribution between high-level abstractions and concepts should be
much easier in the sense of requiring much less data to learn. The hard
part is learning a good representation that does this unfolding
and disentangling.  This may be at the price of a more difficult
training problem, possibly involving ill-conditioning and local minima.

It is only since 2006 that researchers have seriously investigated ways to
train deep architectures, to the exception of the convolutional
networks~\citep{LeCun98-small}. The first realization
(Section~\ref{sec:stacking}) was that unsupervised or supervised layer-wise
training was easier, and that this could be taken advantage of by stacking
single-layer models into deeper ones.

It is interesting to ask {\em why does the layerwise unsupervised pre-training procedure
sometimes help a supervised learner}~\citep{Erhan+al-2010}. There seems to be
a more general principle at play~\footnote{First suggested to us by Leon Bottou}
of {\em guiding the training of intermediate representations}, which may be
easier than trying to learn it all in one go. This is nicely related to the
curriculum learning idea~\citep{Bengio+al-2009-small}, that it may be
much easier to learn simpler concepts first and then build higher-level ones
on top of simpler ones.
This is also coherent with the success of several deep learning algorithms
that provide some such guidance for intermediate representations,
like Semi-Supervised Embedding~\citep{WestonJ2008-small}.

The question of why unsupervised pre-training could be helpful
was extensively studied~\citep{Erhan+al-2010}, trying
to dissect the answer into a {\em regularization effect} and an {\em
  optimization effect}. The regularization effect is clear from the
experiments where the stacked RBMs or denoising auto-encoders are used to
initialize a supervised classification neural
network~\citep{Erhan+al-2010}. It may simply come from the use of
  unsupervised learning to bias the learning dynamics and initialize it in
the {\em basin of attraction} of a ``good'' local minimum (of the training
criterion), where ``good'' is in terms of generalization error.  The
underlying hypothesis exploited by this procedure is that some of the
features or latent factors that are good at capturing the leading
variations in the input distribution are also good at capturing the
variations in the target output random variables of interest (e.g.,
classes). The optimization effect is more difficult to tease out
because the top two layers of a deep neural net can just overfit the
training set whether the lower layers compute useful features or not,
but there are several indications that optimizing the lower levels
with respect to a supervised training criterion can be challenging.

One such indication is that changing the numerical conditions of the
optimization procedure can
have a profound impact on the joint training of a deep architecture,
for example by changing the initialization range and changing the type
of non-linearity used~\citep{GlorotAISTATS2010-small}, much more so
than with shallow architectures. One hypothesis
to explain some of the difficulty in the optimization of deep
architectures is centered on the singular values of the Jacobian
matrix associated with the transformation from the features at one level
into the features at the next level~\citep{GlorotAISTATS2010-small}. 
If these singular values are
all small (less than 1), then the mapping is contractive in every
direction and {\em gradients would vanish} when propagated backwards
through many layers. This is a problem already discussed for
{\em recurrent neural networks}~\citep{Bengio-trnn93}, which can
be seen as very deep networks with shared parameters at each layer,
when unfolded in time. This optimization difficulty has motivated
the exploration of second-order methods for deep architectures
and recurrent networks, in particular Hessian-free second-order
methods~\citep{martens2010hessian-small,Martens+Sutskever-ICML2011-small}.
Unsupervised pre-training has also been proposed to help
training recurrent networks and temporal RBMs~\citep{SutskeverHintonTaylor2009-small},
i.e., at each time step there is a local signal to guide the discovery
of good features to capture in the state variables: model with the current state (as hidden units)
the joint distribution of the previous state and the current input.
Natural gradient~\citep{amari98natural} methods that can be applied to networks with
millions of parameters (i.e. with good scaling properties) have also been 
proposed~\citep{LeRoux+al-tonga-2008-small,Pascanu+Bengio-arxiv2013}. 
\citet{ICML2011Cho_98-small} 
proposes to use adaptive learning rates for RBM training, along with a novel
and interesting idea for a gradient estimator that takes into
account the invariance of the model to flipping hidden unit bits and inverting
signs of corresponding weight vectors. At least one study
indicates that the choice of initialization (to make the Jacobian
of each layer closer to 1 across all its singular values) could
substantially reduce the training difficulty of deep networks~\citep{GlorotAISTATS2010-small}
and this is coherent with the success of the initialization procedure
of Echo State Networks~\citep{Jaeger-2007}, 
as recently studied by~\citet{Sutskever-thesis2012}.
There are also several experimental results~\citep{GlorotAISTATS2010-small,Glorot+al-AI-2011-small,Hinton2010}
showing that the choice of hidden units non-linearity could influence both training and
generalization performance, with particularly interesting results obtained with 
sparse rectifying units~\citep{Jarrett-ICCV2009-small,Hinton2010,Glorot+al-AI-2011-small,Krizhevsky-2012-small}.
An old idea regarding the ill-conditioning issue with neural networks
is that of {\em symmetry breaking}: part of the slowness of convergence
may be due to many units moving together (like sheep) and all trying to
reduce the output error for the same examples. By initializing
with sparse weights~\citep{martens2010hessian-small} or by using
often saturated non-linearities (such as rectifiers
as max-pooling units), gradients only flow along
a few paths, which may help hidden units to specialize more quickly.
Another promising idea
to improve the conditioning of neural network training is to
nullify the average value and slope of each
hidden unit output~\citep{Raiko-2012-small}, and possibly locally
normalize magnitude as well~\citep{Jarrett-ICCV2009-small}.
The debate still rages between using online methods such as stochastic
gradient descent and using second-order methods on large minibatches
(of several thousand examples)~\citep{martens2010hessian-small,Le-ICML2011-small},
with a variant of stochastic gradient descent recently winning an 
optimization challenge~\footnote{\tt\tiny https://sites.google.com/site/nips2011workshop/optimization-challenges}.

Finally, several recent results exploiting {\em large quantities of
  labeled data} suggest that with proper initialization and choice of
non-linearity, very deep purely supervised networks can be trained
successfully without any layerwise
pre-training~\citep{Ciresan-2010,Glorot+al-AI-2011-small,Seide2011,Krizhevsky-2012-small}.
Researchers report than in such conditions, layerwise unsupervised
pre-training brought little or no improvement over pure supervised learning
from scratch when training for long enough.  This reinforces the hypothesis
that unsupervised pre-training acts as a prior, which may be less necessary
when very large quantities of labeled data are available, but begs the
question of why this had not been discovered earlier. The latest results
reported in this respect~\citep{Krizhevsky-2012-small} are particularly interesting
because they allowed to drastically reduce the error rate of object recognition
on a benchmark (the 1000-class ImageNet task) 
where many more traditional computer vision approaches
had been evaluated 
({\tt\scriptsize http://www.image-net.org/challenges/LSVRC/2012/results.html}).
The main techniques that allowed this success include the 
following: {\em efficient GPU training allowing one to train longer} 
(more than 100 million visits of examples),
an aspect first reported by \citet{HonglakL2009-small,Ciresan-2010},
{\em large number of labeled examples}, {\em artificially transformed
  examples} (see Section~\ref{sec:deformations}), 
{\em a large number of tasks} (1000 or 10000 classes for ImageNet), 
{\em convolutional architecture}
with max-pooling (see section~\ref{sec:prior-knowledge} for these latter
two techniques), 
{\em rectifying non-linearities} (discussed above), 
{\em careful initialization} (discussed above),
{\em careful parameter update and adaptive learning rate heuristics},
{\em layerwise feature normalization} (across features),
and a new {\em dropout} trick based on injecting strong binary multiplicative
noise on hidden units. This trick is similar to the
binary noise injection used at each layer of a stack of denoising
auto-encoders. 
Future work is hopefully going to help identify
which of these elements matter most, how to generalize them
across a large variety of tasks and architectures, and
in particular contexts where most examples are unlabeled,
i.e., including an unsupervised component in the training
criterion.

\vsD
\subsection{Joint Training of Deep Boltzmann Machines \note{AC}}
\label{sec:DBM}
\vsA

We now consider the problem of joint training of all layers of a specific
unsupervised model, the Deep Boltzmann Machine (DBM). Whereas much progress
(albeit with many unanswered questions) has been made on jointly training
all the layers of deep architectures using back-propagated gradients (i.e.,
mostly in the supervised setting), much less work has been done on
their purely unsupervised counterpart, e.g. with DBMs\footnote{Joint
training of all the layers of a Deep Belief Net is much more challenging
because of the much harder inference problem involved.}. Note however 
that one could hope that the successful techniques described in the previous
section could be applied to unsupervised learning algorithms.

Like the RBM, the DBM is another
particular subset of the Boltzmann machine family of models where the units
are again arranged in layers. However unlike the RBM, the DBM possesses multiple
layers of hidden units, with units in odd-numbered layers being conditionally independent
given even-numbered layers, and vice-versa.
With respect to the Boltzmann energy function of Eq.~\ref{eq:boltzmann_energy}, the DBM
corresponds to setting $U=0$ and a sparse connectivity structure in both
$V$ and $W$. We can make the structure of the DBM more explicit by specifying
its energy function. For the model with two hidden layers it is given as:

\small
\vsE
\begin{align}
\energy_\theta^{\mathrm{DBM}}(v, h^{(1)}, h^{(2)}; \theta) =& 
    -v^T W h^{(1)} - 
    \left.h^{(1)}\right.^T V h^{(2)} - \nonumber \\
 &   \left.d^{(1)}\right.^T h^{(1)} -
    \left.d^{(2)}\right.^T h^{(2)} -
    b^T v ,
\label{eq:DBM_energy}
\vsD
\end{align}
\normalsize
\vsG

\noindent
with $\theta = \{W,V,d^{(1)},d^{(2)},b\}$. The DBM can also be characterized as
a bipartite graph between two sets of vertices, 
formed by odd and even-numbered
layers (with $v:=h^{(0)}$).

\vsC
\subsubsection{\bf Mean-field approximate inference} 
\vsA

A key point of departure from the RBM is that 
the posterior distribution over the hidden units (given the visibles) is no longer
tractable, due to the interactions between the hidden units. \citet{Salakhutdinov+Hinton-2009-small} resort to a
mean-field approximation to the posterior. Specifically, in the case of a
model with two hidden layers, we wish to approximate $P\left(h^{(1)},h^{(2)} \mid v\right)$ with the factored
distribution $Q_v(h^{(1)},h^{(2)}) =
\prod_{j=1}^{N_1}Q_v\left(h^{(1)}_{j}\right)\
\prod_{i=1}^{N_2}Q_v\left(h^{(2)}_{i}\right)$, such that the KL
divergence $\mathrm{KL} \left( P\left(h^{(1)},h^{(2)} \mid v\right) \Vert
Q_v(h^{1},h^{2}) \right)$ is minimized or equivalently, that a lower
bound to the log likelihood is maximized:

\small
\vsF
\begin{equation}
\log P(v) > \mathcal{L}(Q_{v}) \equiv
\sum_{h^{(1)}}\sum_{h^{(2)}}Q_v(h^{(1)},h^{(2)})\log\left(\frac{P(v,h^{(1)},h^{(2)})}{Q_v(h^{(1)},h^{(2)})}\right)
\label{eq:variational_lower_bound}
\end{equation}
\normalsize

Maximizing this lower-bound with respect to the mean-field distribution
$Q_v(h^{1},h^{2})$ (by setting derivatives to zero)
yields the following mean field update equations:

\vsF
\small
\begin{align}
    \label{eq:mf1}
    \hat{h}^{(1)}_i &\leftarrow \sigmoid\left( 
        \sum_j W_{ji} v_j + \sum_k V_{ik} \hat{h}^{(2)}_k + d^{(1)}_i
        \right) \\
    \label{eq:mf2}
    \hat{h}^{(2)}_k &\leftarrow \sigmoid\left( 
        \sum_i V_{ik} \hat{h}^{(1)}_i + d^{(2)}_k 
        \right)
\vsD
\end{align}
\vsD
\normalsize

Note how the above equations ostensibly look like a {\em fixed point
recurrent neural network}, i.e., with constant input. In the same way
that an RBM can be associated with a simple auto-encoder, the above 
mean-field update equations for the DBM can be associated with a
{\em recurrent auto-encoder}. In that case the training criterion
involves the reconstruction error at the last or at 
consecutive time steps. This type of model has been explored
by~\citet{Savard-master-small} and~\citet{SeungS1998-small} 
and shown to do a better job at denoising than ordinary auto-encoders.

Iterating Eq.~(\ref{eq:mf1}-\ref{eq:mf2}) until convergence yields the
$Q$ parameters of the ``variational positive phase'' of
Eq.~\ref{eq:variational_gradient}:

\vsE
\small
\begin{align}
\mathcal{L}(Q_{v}) =&
    \E{Q_v}{ 
    \log P(v,h^{(1)},h^{(2)}) -
    \log Q_v(h^{(1)},h^{(2)})} \nonumber \\
=& \E{Q_v}{ 
    - \energy_\theta^{\mathrm{DBM}}(v,h^{(1)},h^{(2)}) 
    - \log Q_v(h^{(1)},h^{(2)}) } \nonumber \\
&    - \log Z_\theta \nonumber \\
\label{eq:variational_gradient}
\frac{\partial \mathcal{L}(Q_{v})} {\partial \theta} &= 
    -\E{Q_v}{ 
   \frac{\partial \energy_\theta^{\mathrm{DBM}}(v,h^{(1)},h^{(2)})}{\partial \theta} 
    } \nonumber \\
&\hspace*{4mm}    + \E{P}{
   \frac{\partial \energy_\theta^{\mathrm{DBM}}(v,h^{(1)},h^{(2)})}{\partial \theta}}
\vsD
\end{align}
\normalsize
\vsE

This variational learning procedure leaves the ``negative phase''
untouched, which can thus be estimated through SML or Contrastive Divergence
\citep{Hinton-PoE-2000} as in the RBM case.

\vsC
\subsubsection{\bf Training Deep Boltzmann Machines}
\vsA

The major difference between training a DBM and an RBM
is that instead of maximizing the likelihood directly, we instead choose
parameters to maximize the lower-bound on the likelihood given in
Eq.~\ref{eq:variational_lower_bound}. The SML-based algorithm for maximizing
this lower-bound is as follows:
\vsB
\begin{enumerate}
\item Clamp the visible units to a training example.
\item Iterate over Eq.~(\ref{eq:mf1}-\ref{eq:mf2}) until convergence.
\item Generate negative phase samples $v^-$, $h^{(1)-}$ and $h^{(2)-}$ through SML.
\item Compute $\partial \mathcal{L}(Q_v) \left/ \partial \theta \right.$
    using the values obtained in steps 2-3.
\item Finally, update the model parameters with a step of 
approximate stochastic gradient ascent.
\vsB
\end{enumerate}

While the above procedure appears to be a simple extension of the highly
effective SML scheme for training RBMs, as we demonstrate in
~\citet{Desjardins-arXiv-2012}, this procedure seems vulnerable to falling in poor
local minima which leave many hidden units effectively dead (not significantly
different from its random initialization with small norm). 

The failure of the SML joint training strategy was noted by
\citet{Salakhutdinov+Hinton-2009-small}. As an alternative, they
proposed a greedy layer-wise training strategy.
This procedure consists in pre-training
the layers of the DBM, in much the same way as the Deep Belief Network: i.e.
by stacking RBMs and training each layer to independently model the output of
the previous layer.
A final joint ``fine-tuning'' is done following the above SML-based procedure.

\vsD
\section{Building-In Invariance  \note{PV
    YB AC}}
\vsA
\label{sec:prior-knowledge}

It is well understood that incorporating prior domain knowledge helps
machine learning. Exploring good strategies for doing so is a very important
research avenue.  However, if we are to advance our understanding of core machine
learning principles, it is important that we keep comparisons between
predictors fair and maintain a clear
awareness of the prior domain knowledge used by different
learning algorithms, especially when comparing their performance on
benchmark problems. 
We have so far only presented algorithms that exploited only
generic inductive biases for high dimensional
problems, thus making them potentially applicable
to any high dimensional problem. 
The most prevalent approach to incorporating prior knowledge is to
hand-design better features to feed a generic classifier, and has been used extensively in
computer vision (e.g. \citep{Lowe99}). 
Here, we rather focus on how \emph{basic}
domain knowledge of the input, in particular its topological structure
(e.g. bitmap images having a 2D structure), may be used to \emph{learn}
better features.

\vsD
\subsection{Generating transformed examples}
\label{sec:deformations}
\vsA

Generalization performance is usually 
improved by providing a larger quantity of representative data. This can
be achieved by generating new examples by applying small random deformations
to the original training examples, using deformations
 that are known not to change the target variables
of interest, e.g., an object class is invariant to small 
transformations of images such as translations, rotations, scaling, or shearing.
This old approach~\citep{Baird90-small} has been recently applied with great
success in the work of~\citet{Ciresan-2010} who used an efficient GPU
implementation ($40\times$ speedup) to train a standard but large
deep multilayer Perceptron on deformed MNIST digits. Using both affine and elastic
deformations~\citep{simard-03-small}, with plain old stochastic gradient descent,
they reach a record 0.32\% classification error rate.

\vsD
\subsection{Convolution and pooling \note{PV,AC}}
\label{sec:convol}
\vsA

Another powerful approach is based on even more basic knowledge of merely the \emph{topological
structure} of the input dimensions. By this we mean e.g., the
2D layout of pixels in images or audio spectrograms,
the 3D structure of videos, the 1D sequential structure of text or of
temporal sequences in general. Based on such structure, one can define 
\emph{local receptive fields}~\citep{Hubel+Wiesel-1959}, so that each
low-level feature 
will be computed from only a subset of the input: a neighborhood in the topology (e.g. a
sub-image at a given position).
This topological locality constraint corresponds to a layer having
a very sparse weight matrix with non-zeros only allowed for topologically
local connections. Computing the associated matrix products can
of course be made much more efficient than having to handle a dense matrix,
in addition to the statistical gain from a much smaller number of free parameters.
In domains with such topological structure, similar input patterns 
are likely to appear at different positions, and nearby values (e.g.
consecutive frames or nearby pixels) are likely to have stronger
dependencies that are also important to model the data. In fact these
dependencies can be exploited to {\em discover} the topology~\citep{NIPS2007-925-small}, i.e.
recover a regular grid of pixels out of a set of vectors without
any order information, e.g. after the elements have been arbitrarily
shuffled in the same way for all examples.
Thus a same local feature computation
is likely to be relevant at all translated positions of the receptive
field. Hence the idea of sweeping such a local feature extractor over the
topology: this corresponds to a
\emph{convolution}, and transforms an input into a similarly shaped \emph{feature
  map}. Equivalently to sweeping, this may be seen as static but differently positioned replicated feature
extractors that all share the same parameters.
This is at the heart of convolutional networks~\citep{LeCun89-small,LeCun98-small}
which have been applied both to object recognition
and to image segmentation~\citep{Turaga2010}.
Another hallmark of the convolutional architecture is that values
computed by the same feature detector applied at several neighboring input
locations are then summarized through a pooling operation, typically taking
their max or their sum. This confers the resulting pooled feature layer some
degree of invariance to input translations, and this style of architecture
(alternating selective feature extraction and invariance-creating 
pooling) has been the basis of convolutional networks, the 
Neocognitron~\citep{Fukushima80}
and HMAX~\citep{Riesenhuber+Poggio-1999} models, 
and argued to be the architecture used by mammalian brains
for object recognition~\citep{Riesenhuber+Poggio-1999,Serre-Wolf-2007,DiCarlo-2012}.
The output of a pooling unit will be the same irrespective of where a specific feature is located inside
its pooling region. Empirically the use of pooling seems to contribute
significantly to improved classification accuracy in object classification
tasks~\citep{LeCun98-small,boureau-icml-10-small,boureau-iccv-11-small}. A
successful variant of pooling connected
to sparse coding is L2 pooling~\citep{hyvarinen-book2009,Koray-08-small,Le2010-short},
for which the pool output is the square root of the possibly weighted
sum of squares of filter outputs.
Ideally, we would like to generalize feature-pooling so as to {\em learn
what features should be pooled together}, e.g. as successfully
done in several papers~\citep{Hyvarinen2000-small,Koray-08-small,Le2010-short,Ranzato2010b-short,Courville+al-2011-small,Coates2011c,gregor-nips-11-small}.
In this way, the pool output learns to be {\em invariant} to the
variations captured by the span of the features pooled.


\subsubsection*{\bf Patch-based training}

The simplest approach for learning a convolutional layer in an
{\em unsupervised} fashion is \emph{patch-based training}:
simply feeding a generic unsupervised feature learning algorithm with local patches
extracted at random positions of the inputs. The resulting feature extractor
can then be swiped over the input to produce the convolutional feature
maps. That map may be used as a new input for the next layer, and the operation repeated
to thus learn and stack several layers.
Such an approach was recently used with Independent
Subspace Analysis~\citep{Le-CVPR2011-small} on 3D video blocks, reaching the
state-of-the-art on Hollywood2, UCF, KTH and YouTube action recognition datasets.
Similarly \citep{Coates2011b} compared several feature learners with
patch-based training and reached state-of-the-art results on several classification
benchmarks. Interestingly, in this work performance was almost as good with very simple
k-means clustering as with more sophisticated feature learners. We however
conjecture that this is the case only because patches are rather low dimensional 
(compared to the dimension of a whole image). A large dataset might provide sufficient
coverage of the space of e.g. edges prevalent in $ 6 \times 6 $ patches,
so that a distributed representation is not absolutely necessary. 
Another plausible explanation for this success is that the clusters
identified in each image patch are then {\em pooled} into a histogram
of cluster counts associated with a larger sub-image. Whereas the output
of a regular clustering is a one-hot non-distributed code, this histogram
is itself a distributed representation, and the ``soft'' k-means
\citep{Coates2011b} representation allows not only the nearest filter
but also its neighbors to be active.

\subsubsection*{\bf Convolutional and tiled-convolutional training}

It is possible to directly train large convolutional layers using an
unsupervised criterion. An early approach~\citep{Jain-Seung-08-small} 
trained a standard but deep
convolutional MLP on the task of denoising images, i.e. as a deep,
convolutional, denoising \emph{auto-encoder}.  Convolutional versions of
the RBM or its extensions have also been
developed~\citep{Desjardins-2008-small,HonglakL2009-small,taylor-eccv-10-small}
as well as a \emph{probabilistic max-pooling} operation built into
Convolutional Deep
Networks~\citep{HonglakL2009-small,HonglakLNIPS2009-small,Krizhevsky2010tr-small}.
Other unsupervised feature learning approaches that were adapted to the
convolutional setting include
PSD~\citep{Koray-08-small,koray-nips-10-small,Jarrett-ICCV2009-small,henaff-ismir-11-small},
a convolutional version of sparse coding called deconvolutional
networks~\citep{Zeiler-cvpr2010-small}, Topographic
ICA~\citep{Le2010-short}, and mPoT that~\citet{Kivinen2012-short} applied
to modeling natural textures.  \citet{Gregor+LeCun-2010,Le2010-short} also
demonstrated the technique of tiled-convolution, where parameters are
shared only between feature extractors whose receptive fields are $k$ steps
away (so the ones looking at immediate neighbor locations are not
shared). This allows pooling units to be invariant to more than just
translations, and is a hybrid between convolutional networks and
earlier neural networks with local connections but no weight
sharing~\citep{LeCun-dsbo86-small,LeCun-cp89}.

%

\subsubsection*{\bf Alternatives to pooling}

Alternatively, one can also use explicit knowledge
of the expected invariants expressed mathematically to define 
transformations that are {\em robust} to a known family of 
input deformations, using so-called {\em scattering operators}~\citep{Mallat-2012,Bruna+Mallat-2011},
which can be computed in a way interestingly analogous to deep convolutional networks
and wavelets. Like convolutional networks, the scattering operators alternate two types of operations:
convolution and pooling (as a norm). Unlike convolutional
networks, the proposed approach
keeps at each level all of the information about the input (in a way that can be inverted),
and automatically yields a very sparse (but very high-dimensional) representation.
Another difference is that the filters are not learned but instead set so as to guarantee
that a priori specified invariances are robustly achieved.
Just a few levels were sufficient to achieve impressive results on several benchmark
datasets.

\vsD
\subsection{Temporal coherence and slow features \note{YB}}
\label{sec:slowness}
\vsA

The principle of identifying slowly moving/changing factors in
temporal/spatial data has been investigated by
many~\citep{Becker92,wiskott:2002,hurri+hyvarinen:2003-small,kording2004-small,cadieu+olshausen:2009-small}
as a principle for finding useful representations. In particular this idea has been
applied to image sequences and as an
explanation for why V1 simple and complex cells behave the way they do.  A
good overview can be found in~\citet{hurri+hyvarinen:2003-small,berkes:2005}.


More recently, temporal coherence has been successfully exploited in deep 
architectures to model video~\citep{MobahiCollobertWestonICML2009-small}.
It was also found that temporal coherence discovered visual features similar to those obtained
by ordinary unsupervised feature learning~\citep{Bergstra+Bengio-2009-small},
and a temporal coherence penalty has been {\em combined} with a training
criterion for unsupervised feature learning~\citep{Zou-Ng-Yu-NIPSwkshop2011},
sparse auto-encoders with L1 regularization, in this case, yielding improved
classification performance.

The temporal coherence prior can be expressed in several ways, the simplest
being the squared difference between feature values at times $t$
and $t+1$. Other plausible temporal coherence priors include the following.
First, instead of penalizing the squared change, penalizing the absolute value
(or a similar sparsity penalty)
would state that most of the time the change should be exactly 0, which would
intuitively make sense for the real-life factors that surround us. Second,
one would expect that instead of just being slowly changing, different factors could
be associated with their own different time scale. The specificity of their time scale could thus
become a hint to disentangle explanatory factors. Third, one would expect that
some factors should really be represented by a {\em group of numbers}
(such as $x$, $y$, and $z$ position of some object in space and the
pose parameters of ~\citet{Hinton-transforming-aa-2011-small}) rather than by
a single scalar, and that these groups tend to move together. Structured
sparsity penalties~\citep{Koray-08-small,Jenatton-2009,Bach2011,gregor-nips-11-small}
could be used for this purpose.

\vsD
\subsection{Algorithms to Disentangle Factors of Variation \note{YB-AC}}
\label{sec:disentangling-algorithms}
\vsA

\note{CAN WE MOVE SOME OF THIS MATERIAL IN THE INTRO DISENTANGLING SECTION
OR JUST CUT IT?}



%
The goal of building invariant features is to remove sensitivity of the representation to directions
of variance in the data that are uninformative to the task at hand. However it
is often the case that the goal of feature extraction is the
\emph{disentangling} or separation of
many distinct but informative factors in the data, e.g., in a video of
people: subject identity, action performed, subject pose relative to the camera, etc. 
In this situation, the
methods of generating invariant features, such as feature-pooling, may be inadequate.

The process of building invariant features can be seen as consisting of two steps. First, low-level features are recovered that account
for the data. Second, subsets of these low level features are pooled
together to form higher-level invariant features, exemplified by the pooling
and subsampling layers of convolutional neural networks.
The invariant representation formed by the pooling features offers an
incomplete window on the data as the detailed representation of
the lower-level features is abstracted away in the pooling procedure. While
we would like higher-level features to be more abstract and exhibit greater
invariance, we have little control over what information is lost through
pooling. 
What we really would like is for a particular feature set to be
invariant to the irrelevant features and disentangle the relevant
features. Unfortunately, it is often difficult to determine \emph{a priori}
which set of features will ultimately be relevant to the task at
hand. 

An interesting approach to taking advantage of some of the
factors of variation known to exist in the data is the {\em transforming
auto-encoder}~\citep{Hinton-transforming-aa-2011-small}: instead of a scalar pattern detector
(e.g,. corresponding to the probability of presence of a particular form in the input)
one can think of the features as organized in groups that include both
a pattern detector and {\em pose parameters} that specify attributes of the
detected pattern. In~\citep{Hinton-transforming-aa-2011-small}, what is assumed
a priori is that pairs of examples (or consecutive ones) are observed with
an {\em associated value for the corresponding change in the pose parameters}.
For example, an animal that controls its eyes {\em knows} what changes to
its ocular motor system were applied when going from one image on its retina to the next. In that work, it is also assumed that the pose changes are the same for
all the pattern detectors, and this makes sense for global changes such
as image translation and camera geometry changes. Instead, we would like
to {\em discover} the pose parameters and attributes that should be associated
with each feature detector, without having to specify ahead of time what
they should be, force them to be the same for all features, and having
to necessarily observe the changes in all of the pose parameters or attributes.

The approach taken recently in the Manifold Tangent Classifier, discussed
in section~\ref{sec:leveraging-manifold}, is interesting in this respect.
Without any supervision or prior knowledge, it finds
prominent local \emph{factors of variation} (tangent vectors to the
manifold, extracted from a CAE, interpreted as locally valid input
"deformations").  Higher-level features are subsequently encouraged to be
\emph{invariant} to these factors of variation, so that they must depend on
other characteristics. In a sense this approach is disentangling valid local
deformations along the data manifold from other, more drastic changes,
associated to other factors of variation such as those that affect class
identity.\footnote{The changes that affect class identity might, in input space, actually be of similar
  magnitude to local deformations, but not follow along the manifold,
  i.e. cross
  zones of low density.}

One solution to the problem of information loss that would fit within the
feature-pooling paradigm, is to consider many overlapping pools of
features based on the same low-level feature set. Such a structure would
have the potential to learn a redundant set of invariant features that may
not cause significant loss of information. However it is not obvious what
learning principle could be applied that can ensure that the features are
invariant while maintaining as much information as possible.  While a Deep
Belief Network or a Deep Boltzmann Machine (as discussed in sections
\ref{sec:stacking} and \ref{sec:DBM} respectively) with two hidden layers would,
in principle, be able to preserve information into the ``pooling'' second
hidden layer, there is no guarantee that the second layer features are more
invariant than the ``low-level'' first layer features. However, there is
some empirical evidence that the second layer of the DBN tends to display more
invariance than the first layer \citep{Erhan-vis-techreport-2010}. 

A more principled approach, from the perspective of ensuring a more robust
compact feature representation, can be conceived by reconsidering the disentangling of features
through the lens of its generative equivalent -- feature composition. Since
many unsupervised learning algorithms have a generative
interpretation (or a way to {\em reconstruct} inputs from their
high-level representation), the generative perspective can provide insight into how to
think about disentangling factors. The majority of the models currently used to
construct invariant features have the interpretation that their low-level features
linearly combine to construct the data.\footnote{As an aside, if we are given only the values of
the higher-level pooling features, we cannot accurately recover the data
because we do not know how to apportion credit for the pooling feature
values to the lower-level features. This is simply the generative version
of the consequences of the loss of information caused by pooling.} This is
a fairly rudimentary form of feature composition with significant
limitations. For example, it is not possible to linearly combine a feature with a generic
transformation (such as translation) to generate a
transformed version of the feature. Nor can we even consider a
generic color feature being linearly combined with a gray-scale stimulus
pattern to generate a colored pattern. It would seem that if we are to take the
notion of disentangling seriously we require a richer interaction of
features than that offered by simple linear combinations.

\vsD
\section{Conclusion  \note{YB-AC-PV}}
\vsA


This review of representation learning and deep learning has covered
three major and apparently disconnected approaches: the probabilistic models
(both the directed kind such as sparse coding and the undirected kind such
as Boltzmann machines), the reconstruction-based algorithms related to
auto-encoders, and the geometrically motivated manifold-learning approaches.
Drawing connections between these approaches is currently a very active area of research and is likely to continue to produce models and methods that take advantage of the relative strengths of each paradigm.

{\bf Practical Concerns and Guidelines.}
One of the criticisms addressed to artificial neural networks and deep
learning algorithms is that they have many hyper-parameters and
variants and that exploring their configurations and architectures is an
art. This has motivated an earlier book on the ``Tricks of the
Trade''~\citep{TricksOfTheTrade-small} of which~\citet{LeCun+98backprop-small} is still
relevant for training deep architectures, in particular what concerns
initialization, ill-conditioning and stochastic gradient descent.
A good and more modern compendium of good training practice, particularly adapted to 
training RBMs, is provided in~\citet{Hinton-RBMguide}, while a similar
guide oriented more towards deep neural networks can be found
in~\citet{Bengio-tricks-chapter-2013}, both of which are part of a novel
version of the above book. Recent work on automating hyper-parameter
search~\citep{Bergstra+Bengio-2012-small,Bergstra+al-NIPS2011,Snoek+al-NIPS2012-small} is also making it more convenient, efficient and reproducible.

{\bf Incorporating Generic AI-level Priors.} 
We have covered many high-level generic priors that we believe could 
bring machine learning closer to AI by improving
representation learning. Many of these priors relate to the assumed
existence of multiple underlying factors of variation, whose variations are
in some sense orthogonal to each other. They are expected
to be organized at multiple levels of abstraction, hence the need for deep
architectures, which also have statistical advantages because they allow
to {\em re-use} parameters in a combinatorially efficient way. Only
a few of these factors would typically be relevant for any particular example,
justifying sparsity of representation. These factors are expected to
be related to simple (e.g., linear) dependencies, with subsets of these
explaining different random variables of interest (inputs, tasks)
and varying in structured ways in time and space (temporal and spatial
coherence). We expect future successful applications of representation learning
to refine and increase that list of priors,
and to incorporate most of them instead of focusing on only one.
Research in training criteria that better take these priors into account
are likely to move us closer to the long-term objective of discovering
learning algorithms that can {\em disentangle} the underlying explanatory factors.

{\bf Inference.}  We anticipate that methods based on directly
parametrizing a representation function will incorporate more and more of
the iterative type of computation one finds in the inference procedures of
probabilistic latent-variable models. There is already movement in the
other direction, with probabilistic latent-variable models exploiting
approximate inference mechanisms that are themselves learned (i.e.,
producing a parametric description of the representation function). A major
appeal of probabilistic models is that the semantics of the latent
variables are clear and this allows a clean separation of the problems of
modeling (choose the energy function), inference (estimating $P(h|x)$), and
learning (optimizing the parameters), using generic tools in each case. On
the other hand, doing approximate inference and not taking that approximation into
account explicitly in the approximate optimization for learning could have
detrimental effects, hence the appeal of learning approximate inference.
More fundamentally, there is the question of the multimodality of the
posterior $P(h|x)$. If there are exponentially many probable configurations
of values of the factors $h_i$ that can explain $x$, then we seem to be
stuck with very poor inference, either focusing on a single mode (MAP
inference), assuming some kind of strong factorization (as in variational
inference) or using an MCMC that cannot visit enough modes of $P(h|x)$.
What we propose as food for thought is
the idea of dropping the requirement of an {\em explicit} representation of
the posterior and settle for an {\em implicit} representation that exploits
potential structure in $P(h|x)$ in order to represent it compactly: even
though $P(h|x)$ may have an exponential number of modes, it may be possible
to represent it with a small set of numbers. For example, consider computing a
deterministic feature representation $f(x)$ that implicitly captures the
information about a highly multi-modal $P(h|x)$, in the sense that all the
questions (e.g. making some prediction about some target concept) that can
be asked from $P(h|x)$ can also be answered from $f(x)$.

{\bf Optimization.} Much remains to be done to better understand the
successes and failures of training deep architectures, both in the supervised
case (with many recent successes) and the unsupervised case (where much
more work needs to be done). Although regularization effects can be important
on small datasets, the effects that persist on very large datasets suggest
some optimization issues are involved. Are they more due to local minima
(we now know there are huge numbers of them) and the dynamics of the
training procedure? Or are they due mostly to ill-conditioning and may be handled
by approximate second-order methods? These basic questions remain unanswered
and deserve much more study.

\vsC
\subsubsection*{\bf Acknowledgments}
\vsA

The author would like to thank David Warde-Farley, Razvan Pascanu and
Ian Goodfellow for useful feedback, as well as NSERC, CIFAR and the Canada Research
Chairs for funding.

\vsD
{\small
\bibliographystyle{natbib}
\bibliography{strings,ml,aigaion}
}


\ifarxiv
\else
\begin{IEEEbiography}
[{\includegraphics[width=1in,height=1.25in,clip,keepaspectratio]{yoshua.jpg}}]
{Yoshua Bengio}
is Full Professor of the Department of Computer Science and Operations
Research and head of the Machine Learning Laboratory (LISA) at the
University of Montreal, CIFAR Fellow in the Neural Computation and Adaptive
Perception program, Canada Research Chair in Statistical Learning
Algorithms, and he also holds the NSERC-Ubisoft industrial chair. His main
research ambition is to understand principles of learning that yield
intelligence.
\end{IEEEbiography}


\begin{IEEEbiography}
[{\includegraphics[width=1in,height=1.25in,clip,keepaspectratio]{aaron.jpg}}]
{Aaron Courville}
is an Assistant Professor of the Department of Computer Science and
Operations Research at the University of Montreal. His recent research
interests have focused on the development of deep learning models and
methods. He is particularly interested in developing probabilistic models
and novel inference methods.
\end{IEEEbiography}


\begin{IEEEbiography}
[{\includegraphics[width=1in,height=1.25in,clip,keepaspectratio]{pascal.jpg}}]
{Pascal Vincent}
is an Associate Professor of the Department of Computer Science and
Operations Research at the University of Montreal, and CIFAR Associate in
the Neural Computation and Adaptive Perception program.  His recent
research has focused on novel principles for learning direct encodings of
representations.  He is also interested in developing alternative parameter
estimation techniques for probabilistic models.
\end{IEEEbiography}
\fi
\end{document}